\documentclass[10pt,journal,compsoc]{IEEEtran}
\ifCLASSOPTIONcompsoc
  \usepackage[nocompress]{cite}
\else
  \usepackage{cite}
\fi
\ifCLASSINFOpdf
\else
\fi
\usepackage{graphicx}
\usepackage{amsmath}
\usepackage{amssymb}
\usepackage{dirtytalk}
\usepackage{bbm}
\usepackage{derivative}
\usepackage[pagebackref,breaklinks,colorlinks]{hyperref}
\usepackage{xcolor}
\hypersetup{
    colorlinks,
    linkcolor={black},
    citecolor={black},
    urlcolor={black}
}
\usepackage{algorithm,algpseudocode}
\usepackage{mathtools}
\DeclarePairedDelimiterX{\norm}[1]{\lVert}{\rVert}{#1}
\usepackage[hang]{footmisc}
\usepackage{tabularx,ragged2e}
\usepackage{multirow}
\usepackage{booktabs}

\newcolumntype{C}{>{\Centering\arraybackslash}X}  
\newcolumntype{t}{>{\Centering\hsize=.6\hsize}X}
\newcolumntype{s}{>{\Centering\hsize=.5\hsize}X}
\newcolumntype{j}{>{\Centering\hsize=.16\hsize}X}
\newcolumntype{k}{>{\Centering\hsize=.3\hsize}X}
\newcolumntype{y}{>{\Centering\hsize=.12\hsize}X}
\newcolumntype{i}{>{\Centering\hsize=.08\hsize}X}
\newcolumntype{v}{>{\Centering\hsize=.04\hsize}X}
\algdef{SE}[SUBALG]{Indent}{EndIndent}{}{\algorithmicend\ }%
\algtext*{Indent}
\algtext*{EndIndent}
\newcommand{\op}[1]{\operatorname{#1}}
\usepackage{tikz}

\usepackage{textcomp}
\newcommand\copyrighttext{%
\footnotesize \textcopyright \hspace{0.1mm} 2023 IEEE.  Personal use of this material is permitted.  Permission from IEEE must be obtained for all other uses, in any current or future media, including reprinting/republishing this material for advertising or promotional purposes, creating new collective works, for resale or redistribution to servers or lists, or reuse of any copyrighted component of this work in other works. DOI: \href{https://doi.org/10.1109/TPAMI.2023.3243812}{10.1109/TPAMI.2023.3243812}
}
\newcommand\copyrightnotice{%
\begin{tikzpicture}[remember picture,overlay]
\node[anchor=south, yshift=4pt] at (current page.south) {\fbox{\parbox{\dimexpr\textwidth-\fboxsep-\fboxrule\relax}{\copyrighttext}}};
\end{tikzpicture}%
}

\begin{document}
\title{Self-supervised Video-centralised Transformer \\ for Video Face Clustering}

\author{
Yujiang Wang,  
Mingzhi Dong,
Jie Shen, \IEEEmembership{Member, IEEE,}
Yiming Luo, 
Yiming Lin, \IEEEmembership{Member, IEEE,}
Pingchuan Ma, \IEEEmembership{Member, IEEE,}
Stavros Petridis, \IEEEmembership{Member, IEEE,}
and Maja Pantic, \IEEEmembership{Fellow, IEEE}
\IEEEcompsocitemizethanks{
\IEEEcompsocthanksitem{Yujiang Wang is with Department of Computing, Imperial College London, UK (e-mail: {yujiang.wang14@imperial.ac.uk}).}
\IEEEcompsocthanksitem{Mingzhi Dong is with School of Computer Science, Fudan University, China (e-mail: {mingzhidong@gmail.com}).}
\IEEEcompsocthanksitem{Jie Shen, Yiming Luo, Yiming Lin, Pingchuan Ma, Stavros Petridis, and Maja Pantic are with Department of Computing, Imperial College London, UK (e-mail: 
{jie.shen07@imperial.ac.uk}
{yiming.luo22@imperial.ac.uk};
{yiming.lin15@imperial.ac.uk}; 
{pingchuan.ma16@imperial.ac.uk};
{stavros.petridis04@imperial.ac.uk};
{maja.pantic@gmail.com}).
}
\IEEEcompsocthanksitem{Jie Shen is the corresponding author (e-mail: jie.shen07@imperial.ac.uk).}}
}


\IEEEtitleabstractindextext{
\begin{abstract}
This paper presents a novel method for face clustering in videos using a video-centralised transformer. Previous works often employed contrastive learning to learn frame-level representation and used average pooling to aggregate the features along the temporal dimension. This approach may not fully capture the complicated video dynamics. In addition, despite the recent progress in video-based contrastive learning, few have attempted to learn a self-supervised clustering-friendly face representation that benefits the video face clustering task. To overcome these limitations, our method employs a transformer to directly learn video-level representations that can better reflect the temporally-varying property of faces in videos, while we also propose a video-centralised self-supervised framework to train the transformer model. We also investigate face clustering in egocentric videos, a fast-emerging field that has not been studied yet in works related to face clustering. To this end, we present and release the first large-scale egocentric video face clustering dataset named EasyCom-Clustering. We evaluate our proposed method on both the widely used Big Bang Theory (BBT) dataset and the new EasyCom-Clustering dataset. Results show the performance of our video-centralised transformer has surpassed all previous state-of-the-art methods on both benchmarks, exhibiting a self-attentive understanding of face videos. 
\end{abstract}

}

\maketitle
\copyrightnotice
\IEEEdisplaynontitleabstractindextext
\IEEEpeerreviewmaketitle

\IEEEraisesectionheading{\section{Introduction}\label{sec:introduction}}
\IEEEPARstart{T}{he} idea of video face clustering can be traced back to the pursuit of automatic labelling of character names in TV shows \cite{everingham2006hello}.
Aiming to classify facial images detected in videos into identity-based clusters without any supervision, video face clustering has become increasingly influential and valuable today. The prevalence of easy-to-access video streaming platforms has led to rapid growth in the amount of public video data that are most relevant to human activities and behaviours.
The character-level analysis is essential to understanding those videos, while a successful video face clustering framework can answer the key question of character understanding, \textit{Who are those people?} without the involvement of any laborious human annotation. This feature can also be extremely useful for other applications such as video captioning \cite{yan2019stat}, video summarisation \cite{ji2019video}, content-based video retrieval \cite{spolaor2020systematic}, etc.

Despite the broad applications, video face clustering is still a challenging task. Facial images in real-world videos are typically accomplished with a high level of intra-class variations, e.g. the varying facial appearance due to head pose changes, varying illuminations, different backgrounds, facial expressions, occlusions, and so on. Those facial images that are more difficult to be classified are therefore named as \textit{visual distractors}  \cite{somandepalli2020multi}, and how to appropriately cluster those visual distractors stands at the heart of video face clustering. Further challenges can arise from the continuation of videos, i.e. the newly detected faces may or may not be seen previously, and certain characters may only be observed once. 

Most relevant works \cite{sharma2019self, Kalogeiton2020bmcv,sharma2020clustering,somandepalli2020multi} address those challenges through the utilisation of \textit{must-link} and \textit{cannot-link} constraints reasoned automatically from face detection. Unlike face recognition in still images, video face clustering can benefit from the temporal property of a video sequence, based on one simple yet rather effective heuristic. That is, the temporally consecutive facial images with overlapped detection boxes are deemed as from the same person, \cite{xiang2015learning, wu2013simultaneous, wu2013constrained} and they can be grouped together to formulate a \textit{face track} \cite{sivic2005person}. All facial images within a face track can be regarded to have the same identity, which is the so-called \textit{must-link}. If two face tracks have co-occurred, it is safe to conclude that they are from different persons, given that the two subjects have appeared in the same frame simultaneously. This is the \textit{cannot-link} constraint. It should be noticed that both must-links and cannot-links can be automatically gathered from face detection in videos, and video face clustering can be considered to be a self-supervised problem with such prior knowledge embedded. \cite{xiao2014weighted} 

Most prior works of video face clustering, therefore, can be roughly divided into two categories based on how they utilise the constraints. The first category of works focuses on the improvement of the clustering stage \cite{somandepalli2020multi, Kalogeiton2020bmcv, cao2015constrained, wu2013constrained, antonopoulos2007hierarchical}. Facial embedding for each target facial image is extracted using a certain feature extractor, and the main interest is to better cluster those descriptors with must-links and cannot-links. Some representative solutions include the application of constrained multi-view clustering \cite{cao2015constrained, somandepalli2020multi},  constrained 1NN clustering \cite{Kalogeiton2020bmcv}, Erdos-Renyi clustering \cite{jin2017end}, Hidden Markov Random Fields \cite{wu2013constrained}, etc. Those works usually do not consider how to learn a better facial representation using constraints. 

Another family of video face clustering concentrates on the generation of more discriminative and robust facial representations. Despite a few earlier attempts  \cite{xiao2014weighted}, this route started to receive attention from the community since deep learning has been popular. A basic video face clustering framework is proposed in \cite{Sharma2017ASimple}. A pre-trained CNN model is first involved to extract frame-level face descriptors, and then a simple temporal average operator is applied to summarise the descriptors into track-level ones. The clustering is then performed on those track-level representations using Hierarchical Agglomerative Clustering (HAC). In TSiam \cite{sharma2019self}, this framework is further extended through the application of contrastive learning \cite{hadsell2006dimensionality} to train a Siamese Multi-Layer Perceptron (MLP) that can generate more discriminative facial representations. The general idea is that all pairs of descriptors selected from the same face track (must-link) are positive samples, while the pairs drawn from two co-occurring face tracks (cannot-link) can be seen as negative ones. A Siamese MLP can therefore be trained with contrastive learning to project facial descriptors into a new latent space, followed by a temporal aggregation and a clustering algorithm like HAC. The pipeline of TSiam is inspiring and can be found in other video face clustering methods like SSiam \cite{sharma2019self} and CCL \cite{sharma2020clustering}. The main difference lies in the way of generating contrastive supervisions, i.e. instead of using must-links and cannot-links, SSiam \cite{sharma2019self} and CCL \cite{sharma2020clustering} explore how to determine positive/negative samples via mining challenging data and via the usage of FINCH \cite{sarfraz2019efficient} clustering algorithm, respectively. In those works, Siamese MLP is a popular choice for obtaining more discriminative representations, and its effectiveness has been validated by the superior clustering performance achieved. 

However, the temporal nature of video face clustering still leaves room to improve the Siamese MLP-based framework. As noticed in \cite{dave2021tclr}, videos consist of both temporally-invariant and temporally-varying properties. 
In the domain of video face clustering, 
those properties can be intuitively understood as the absences and presences of visual distractors, respectively. When a face track only consists of facial images that can be distinguished with ease, for instance, a track of constantly frontal faces without significant changes, it can be illustrated as a temporally-invariant video. However, face clustering in real-world videos will inevitably introduce visual distractors that can be confusing to classifiers, and it is common to find face tracks with multiple visual distractors. As such, the temporally-varying aspects of the face track should be more explicitly considered to increase the robustness against visual distractors.
The Siamese MLP-based methods, however, follow a temporally-invariant way of constructing track-level representations, i.e. a simple temporal pooling like average, and thus they fail to reflect the temporally-varying attributes in face tracks, which may lead to degraded video clustering performance. 

In this paper, unlike previous works that learn a frame-level representation and then perform temporally-invariant aggregation, we propose to directly obtain a video-level representation for each face track via a transformer model \cite{vaswani2017attention}. Characterised by the self-attention mechanisms, transformers can implicitly capture the global relationships within the data and dynamically adjust the weights of each input signal, and therefore it can be a promising architecture to learn the temporally-variant and temporally-varying properties in videos. Through the usage of the transformer, we are aiming to achieve a video-level understanding of different persons in face videos. 

Learning such a transformer, however, is not a straightforward task, as video face clustering is a self-supervised task that requires clustering-friendly representations. The transformer is first proposed for Natural Language Processing (NLP) tasks \cite{vaswani2017attention}. The recent progress of Vision Transformer \cite{dosovitskiy2020image} (ViT) has indicated that it can also be applied to visual recognition tasks and can achieve competitive performance against the commonly-used Convolutional Networks (ConvNets). Inspired by ViT, the application of transformers in image-based computer vision tasks has received extensive research interest \cite{liu2021swin, touvron2021training, zhou2021deepvit, touvron2021going, yuan2021tokens, chen2021crossvit}.
There are also some works on video-based transformers \cite{bertasius2021space, arnab2021vivit,  zhang2021vidtr}. However, self-supervised video understanding with transformers is still limited
\cite{wang2021long}.
To the best of our knowledge, there are still no studies on how to train a self-supervised video-based transformer that can better benefit video clustering tasks.

\begin{figure*}[ht!]
\centering
\includegraphics[width=0.99\linewidth]{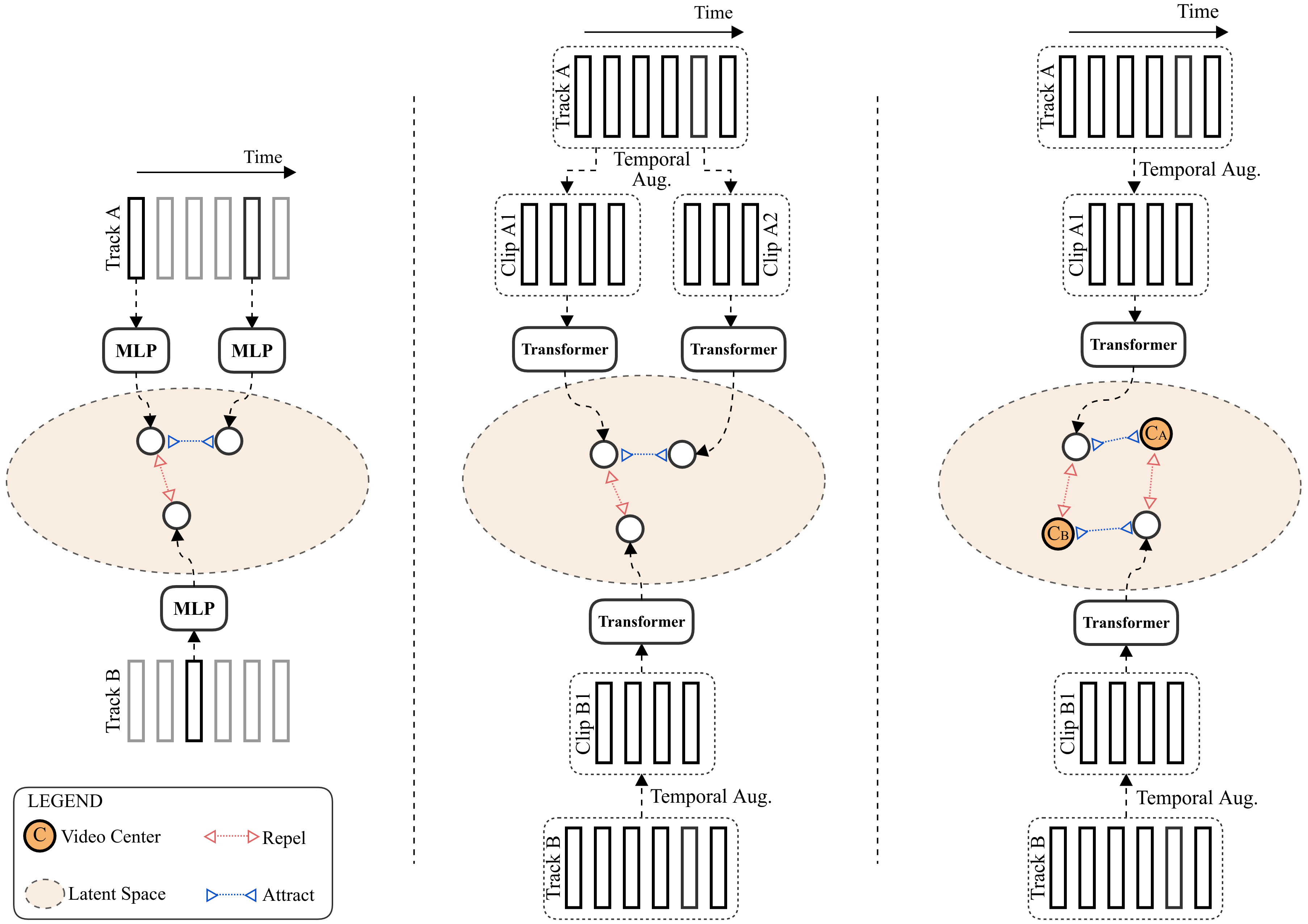} 
\caption{Illustrations of the learning paradigms of different video face clustering methods. \protect\say{Track A} and \protect\say{Track B} are the two face tracks of descriptors that are known to be from a different person. \hspace{1mm} \textit{Left}: Previous works on video face clustering mainly utilised contrastive learning to train a frame-level Siamese MLP.  \hspace{1mm} \textit{Middle}: A naive contrastive learning framework of using a transformer model to obtain video-level representation. Temporal augmentation is involved to sample clips from the face track, while the contrastive learning is still performed in a pairwise manner.  \hspace{1mm}  \textit{Right}: Our proposed video-centralised transformer. Each face tack is assigned a distinct video centre, and the objective is to attract the video representation of each sampled clip to its own centre, while repelling its distances with centres of those cannot-link face tracks. (Best seen in colour)}
\label{fig: idea_explain}
\end{figure*}

Inspired by the idea of deep clustering \cite{xie2016unsupervised, yang2017towards}, we propose the video-centralised learning to train a self-supervised transformer for video face clustering, based on the intuitive motivation that each face track should maintain a distinct video centre in the latent space. The general learning paradigms between the previous works and ours are illustrated in Fig. \ref{fig: idea_explain}. As shown in Fig. \ref{fig: idea_explain} (\textit{Left}), contrastive learning \cite{hadsell2006dimensionality} is utilised in \cite{sharma2019self,sharma2020clustering} to train a Siamese MLP on frame-level facial descriptors.  
The objective is to minimise the distance between the two positive descriptors in the new latent space and to maximise the distance between negative ones. 
A naive extension of this learning paradigm to transformer models is displayed in Fig. \ref{fig: idea_explain} (\textit{Middle}).
The major modification is that the new representation is obtained via a transformer with temporally-augmented clips as input, while the contrastive learning is still conducted in a pairwise way as in Fig. \ref{fig: idea_explain} (\textit{Left}).  
Note this paradigm of training a transformer has not yet been adopted by any other video face clustering works.

Our proposed video-centralised learning is depicted in Fig. \ref{fig: idea_explain} (\textit{Right}). For each face track, we maintain a distinct video centre in the new latent space. 
When sampling a clip from a face track, the latent representation predicted by a transformer should be attracted to the video centre of that track, and also should be repelled from the centres of tracks that are already known to be negative, e.g. cannot links. With the introduction of video centres, we are enforcing the video representation of the same face track to be more compact and to be more \textit{centralised},
which can lead to a more clustering-friendly structure in the latent space. This kind of representation is generally more desirable for a clustering problem like video face clustering. Note that there is no need to pair-wisely sample the data in our video-centralised learning, given the video centres can be learned jointly with the transformer.
How to define and learn video centres is a crucial problem here, and we have explored several potential options to determine the best-working one in this work. 

As an additional contribution, we also investigate video face clustering in egocentric videos, given that most previous works are evaluated on datasets of TV shows or movies. Compared with those videos taken by third-person cameras, egocentric videos \cite{li2013learning, lu2013story, lee2012discovering} are usually recorded by first-view wearable devices. Character analysis in egocentric videos is becoming increasingly crucial today with the rapid growth of live video streaming platforms and VR/AR markets. To facilitate the research on egocentric video face clustering, we extract the facial embedding for all the $94\,047$ face tracks of the recently released EasyCom dataset \cite{donley2021easycom} using ArcFace \cite{deng2019arcface}, and annotate each track with human-examined identities. The extracted face tracks and annotations constitute a complementary dataset of EasyCom \cite{donley2021easycom} in the field of video face clustering, and it is named EasyCom-Clustering\footnote{Available at \url{https://github.com/ibug-group/Easycom-Clustering}}. EasyCom-Clustering is a large-scale dataset and significantly exceeds the commonly used TV show datasets like \textit{Big Bang Theory} (BBT) \cite{roy2014tvd}, \textit{Buffy the Vampire Slayer} \cite{everingham2006hello} in terms of annotated face tracks and total duration.

We mainly compare the performance of the proposed video-centralised transformer with that of three state-of-the-art video face clustering methods, TSiam \cite{sharma2019self}, SSiam \cite{sharma2019self}, CCL \cite{sharma2020clustering}, plus a Siamese contrastive transformer (Fig. \ref{fig: idea_explain} \textit{Middle}). All approaches are evaluated on two datasets of different video domains, i.e. 1). BBT dataset \cite{roy2014tvd, tapaswi2019video} of $6$ episodes representing TV shows, and 2). our EasyCom-Clustering dataset consisting of 22 egocentric videos. The proposed method has outperformed all baselines on those two datasets under two different clustering scenarios, and it also has exhibited a certain degree of video understanding which is not achievable using baseline methods. 

Overall, we have made the following contributions in this paper:

\begin{itemize}
\item We propose to tackle video face clustering from a video-level perspective to better model the temporally-varying property in face videos. A self-supervised transformer \cite{vaswani2017attention} is employed to address this issue, which is the first time to the best of our knowledge.

\item A self-supervised video-centralised learning that can generate more clustering-friendly representations is proposed to train the transformer model. The performance achieved on two datasets has demonstrated the effectiveness of the proposed video-centralised transformer. 

\item We also release a large-scale dataset, EasyCom-Clustering, to facilitate the research on egocentric video face clustering. A benchmark evaluation of different methods on this dataset is also provided. 

\end{itemize}

\section{Related Works}

\subsection{Video Face Clustering}
Face clustering is a computer vision problem with a long view of history \cite{ho2003clustering, berg2004names}. Research on image-based face clustering \cite{he2018merge,zhu2011rank,otto2017clustering,lin2018deep} aims to reveal the underlying relationships between facial image representations through the usage of Approximate Nearest Neighbours \cite{otto2017clustering, shi2018face}, the graph convolution networks (GCN) \cite{wang2019linkage, yang2019learning}, unsupervised clustering methods \cite{lin2018deep, lin2017proximity}, etc. Face clustering in images is commonly considered an unsupervised problem, as it typically employs a pre-trained CNN to extract the facial descriptors to be clustered without supervision.

Face clustering in videos is different from image-based face clustering due to the availability of must-link and cannot-link constraints. Given a video with face detection, the two constraints can be reliably exploited using the simple and effective heuristic of consecutive face detection overlapping \cite{sivic2005person, wu2013simultaneous, wu2013constrained}. Therefore, the major concern in video face clustering is how to incorporate those constraints,  and therefore video face clustering can be regarded as a self-supervised problem. Based on how must-links and cannot-links are used, the works can be roughly categorised into two groups, constrained clustering methods \cite{wu2013constrained, cao2015constrained, Kalogeiton2020bmcv} and contrastive learning approaches \cite{sharma2019self, sharma2020clustering, tapaswi2019video}. The first group focuses on improving the clustering algorithm in latent space with constraints. Treating face tracks as multi-view data, some works adopt Constrained Multi-View Spectral Clustering \cite{cao2015constrained} or Multi-view Correlation Objective \cite{somandepalli2020multi} as solutions. An Erdos-Renyi clustering method is proposed by Jin et al. \cite{jin2017end} based on the idea of ranking counts. Hidden Markov Random Fields is also a popular choice to model such constraints in the clustering space \cite{wu2013constrained, wu2013simultaneous}.
Recently, Kalogeiton and Zisserman \cite{Kalogeiton2020bmcv} improve the FINCH clustering \cite{sarfraz2019efficient} with must-link and cannot link constraints, leading to a Constrained 1NN clustering method. These works usually use a pre-trained feature extractor to obtain facial embedding, without further investigation of the facial representations.  

Another group of video face clustering focuses on obtaining more discriminative and robust facial representations. The Weighted Block-Sparse Low-Rank Representation is proposed in \cite{xiao2014weighted} to learn a low-rank facial representation with prior knowledge from constraints. A simple video face clustering framework is presented by Sharma et al. \cite{Sharma2017ASimple}, using a pre-trained CNN to extract facial descriptors and performing clustering at temporally aggregated track representations. This framework is further advanced by TSiam and SSiam \cite{sharma2019self} based on the idea of contrastive learning \cite{hadsell2006dimensionality}. A Siamese MLP is adopted to learn a more discriminate facial representation from either track-level supervision (TSiam) or the mining of challenging samples (SSiam). Similar ideas can be found in CCL \cite{sharma2020clustering} that utilise weak supervisions generated by FINCH algorithm \cite{sarfraz2019efficient} to train the Siamese MLP. Video face clustering with unknown cluster number is studied in \cite{tapaswi2019video}, while the authors of \cite{zhang2016deep} developed an improved triplet loss based on deep metric learning. The Siamese MLP-based methods like TSiam \cite{sharma2019self} and CCL \cite{sharma2020clustering} have achieved impressing clustering performance in video face clustering datasets such as \textit{Big Bang Theory} \cite{roy2014tvd} or \textit{Buffy the Vampire Slayer} \cite{everingham2006hello}. However, the lack of video-level understanding may impair their potential. In this work, we examine how to obtain video-level facial representation via transformer \cite{vaswani2017attention}. 

Besides, the datasets used by previous works are mostly in a specific video domain, i.e. the TV shows taken with third-person cameras, while the analysis of egocentric videos \cite{liu20214d, huang2018predicting, ng2020you2me} is increasingly receiving attention with the popularity of first-person live video streaming platform, VR/AR devices, etc. However, to the best of our knowledge, there is still no egocentric-based video face clustering dataset so far. In this paper, we present the first large-scale egocentric video face clustering dataset, EasyCom-Clustering, to facilitate relevant research, and we also evaluate multiple methods including ours on it as benchmarks. EasyCom-Clustering is a complementary dataset on clustering of the recently released EasyCom \cite{donley2021easycom}. 

\subsection{Self-supervised Video Understanding with Transformers}
Transformers \cite{vaswani2017attention} are initially developed for Natural Language Processing (NLP) tasks. The introduction of the multi-head attention mechanism enables transformers to capture the global dependency in a self-attentive manner. But the performance of transformers on visual tasks was not very satisfying until the pioneering work of Vision Transformers (ViT) \cite{dosovitskiy2020image}. ViT is a variant of the transformer model that is modified to address general image recognition tasks, and it has achieved 
very competitive performance against the commonly-used ConvNets. Since ViT, image recognition with transformers has attracted enormous attention, and considerable efforts have been made to improve ViT architectures, including Swin transformer with local attentions \cite{liu2021swin}, distillation transformer \cite{touvron2021training}, Deeper ViT \cite{zhou2021deepvit}, Token-to-Token ViT \cite{yuan2021tokens}, just to name a few of them. We refer the readers to \cite{khan2021transformers} for more details. 

The application of transformers in the video domain has also been widely investigated, mainly via extending the 2D-based self-attention mechanism into spatial-temporal ones \cite{bertasius2021space, zhang2021vidtr, akbari2021vatt}. TimeSformer \cite{bertasius2021space} aims to increase the computational efficiency of transformers on spatial-temporal dimensions via factorisation. VidTr \cite{zhang2021vidtr} shares a similar target of reducing computational complexity and memory usage of video-based transformers through a standard-deviation-based topK pooling. Swin transformer \cite{liu2021swin} for image recognition tasks is extended into a Space-Time Swin transformer in \cite{wang2021long} by leveraging 3D shifting windows. 
Fan et al. \cite{fan2021multiscale} propose a multi-scale ViT that can efficiently operate on videos. However, since video face clustering is a self-supervised problem, our work is more related to self-supervised video understanding. 

Recent works on self-supervised video understanding \cite{oord2018representation, feichtenhofer2021large, han2020self, han2019video,wang2021long, qian2021spatiotemporal} usually follow the popular self-supervised contrastive learning framework like 
MoCo \cite{he2020momentum}, SimCLR \cite{chen2020simple}, etc.  
The general idea is that the video clips sampled (usually with several temporal/spatial augmentations) from the same video are deemed as positive samples, and an InfoNCE loss \cite{oord2018representation} which is essentially a cross-entropy loss is adopted to maximise the probability distributions of those positive samples. In this field, however, the application of video-based transformer is comparatively rare, while the Long-Short Temporal Contrastive Learning (LSTCL) \cite{wang2021long} is the most related one, to the best of our knowledge. In LSTCL \cite{wang2021long}, a long and a short clip are sampled simultaneously to train a video transformer on several popular self-supervised contrastive learning frameworks. 
None of those works has ever explored the generation of self-supervised clustering-friendly video representations through transformers. 
In this work, we take inspiration from deep clustering \cite{xie2016unsupervised, yang2017towards} to train such a transformer.

\subsection{Deep Clustering}
Research on deep clustering  \cite{yang2016joint, xie2016unsupervised, guo2017deep, yang2017towards} typically involves a clustering-guided method to optimise the feature representations, and the assignment of cluster centroids is a crucial idea. 
There are also some relevant works leveraging the predictions from clustering algorithms as pseudo labels to perform unsupervised learning such as \cite{caron2018deep}, but we focus on centroid-based representations.
Deep Embedded Clustering \cite{xie2016unsupervised} employs Student's T-distribution as a soft-assignment to measure the similarity between a latent representation and cluster centroids, which allows the gradients to be back-propagated to a deep network. Similar ideas can be found in \cite{yang2017towards}. It proposes to optimise cluster centroids and latent representations iteratively to obtain a k-means-friendly representation. The determination of the total cluster number is usually a pre-requisite in those methods, while our video-centralised framework assigns a distinct video centre for each face track, with no need to estimate the cluster number. Our work is also distinguished from others in that we design a clustering-friendly self-supervised framework for video understanding with transformers for the first time. 

\begin{figure*}[ht!]
\centering
\includegraphics[width=0.99\linewidth]{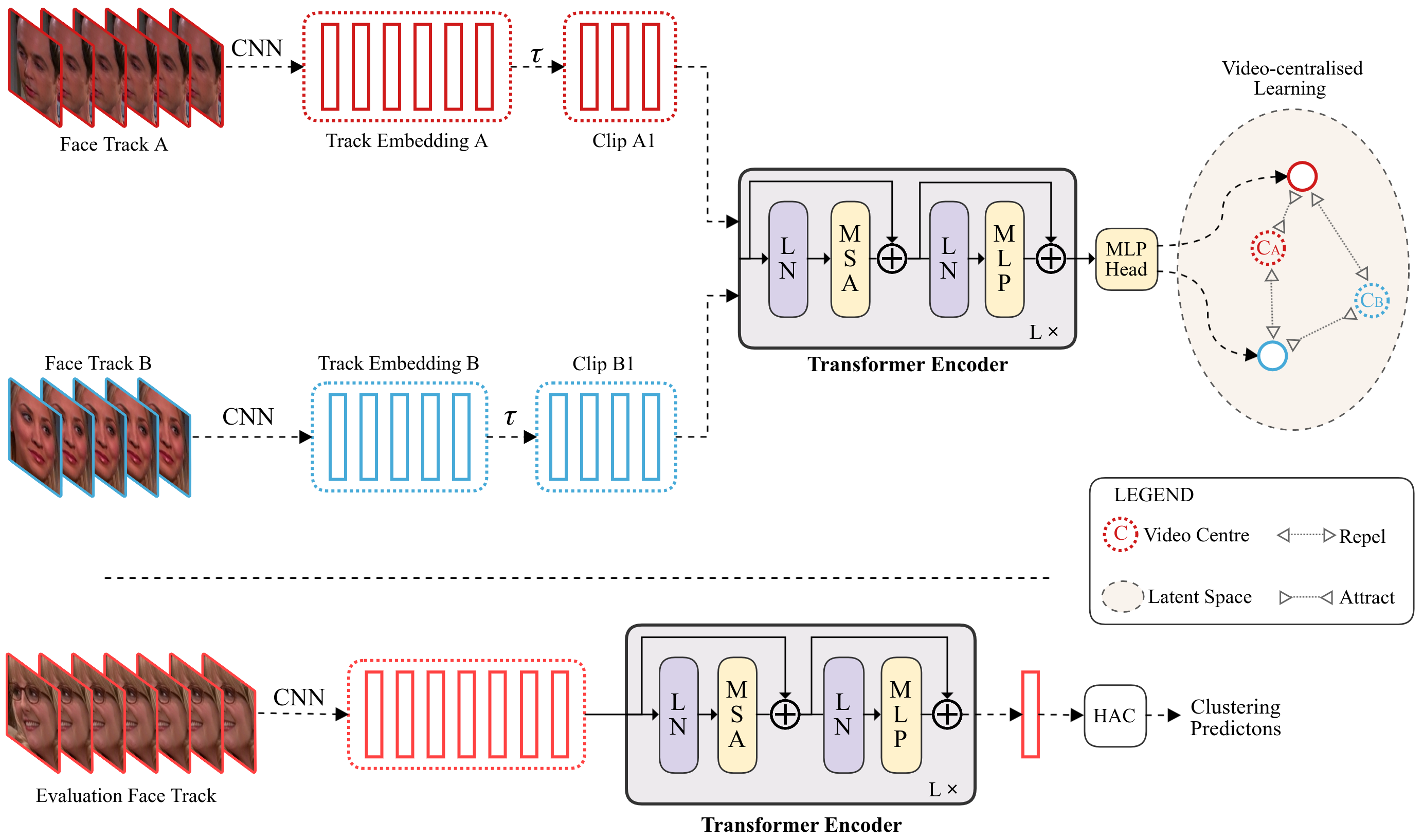} 
\caption{An illustration of the general training and evaluation framework of the proposed video-centralised transformer. \hspace{1mm} \textit{Top}: The training framework. For each face track, a ConvNet (CNN) is first employed to extract embedding, followed by a temporal augmentation $\op{\tau}$. The sampled clip is fed into a transformer encoder with an MLP head to a new latent space with reduced dimensionality. The video-centralised learning is performed in this new space. \hspace{1mm} \textit{Bottom}: The evaluation pipeline. The embedding of each face track is also extracted first, while the whole track is input into the transformer encoder without augmentation $\op{\tau}$. The output from the encoder is used as the video representation, while the MLP head is also discarded. The final clustering results are obtained by a HAC clustering algorithm.
(Best seen in colour)}
\label{fig: framework}
\end{figure*}

\section{Video-centralised Transformer}
\label{sec:method}

\subsection{Problem Definition and Backgrounds}
With prior knowledge such as must-links (a face track itself) and cannot-links (two temporally co-occurring face tracks) embedded, video face clustering is essentially a self-supervised problem. 
For a given video $\mathcal{V}$, let $\mathcal{T}_m=\{\mathbf{x}_1^m, \mathbf{x}_2^m, ..., \mathbf{x}_n^m\}$ be the $m$-th face track of length $n$ where $\mathbf{x}_i^m$ denotes the $i$-th facial image, and assuming there are a total of $M$ face tracks. We can denote the video as the collections of those tracks for simplicity, i.e. $\mathcal{V}=\{\mathcal{T}_1, \mathcal{T}_2, ..., \mathcal{T}_M\}$, and the target is to classify those face tracks into $K$ disjoint clusters of different identities (IDs). 

Two constraints can be leveraged to address this issue, i.e. the must-link and cannot-link, respectively. The must-link indicates that two arbitrary facial images drawn from the same track can be deemed to have a common ID. This constraint can be written as a set $\mathcal{M}=\{(\mathbf{x}_i^m, \mathbf{x}_j^m) \mid \forall \mathbf{x}_i^m,\mathbf{x}_j^m \in \mathcal{T}_m, i \neq j, \forall \mathcal{T}_m \in \mathcal{V}\}$. Let $\mathbbm{1}(\mathbf{x}_i, \mathbf{x}_j)$ be the indicator function for identity, i.e. $\mathbbm{1}(\mathbf{x}_i, \mathbf{x}_j)=1$ if $\mathbf{x}_i$ and $\mathbf{x}_j$ have the same ID, and $\mathbbm{1}(\mathbf{x}_i, \mathbf{x}_j)=0$ otherwise. We have $\mathbbm{1}(\mathbf{x}_i, \mathbf{x}_j)=1, \forall (\mathbf{x}_i,\mathbf{x}_j) \in \mathcal{M}$. Cannot-links, on the other hand, are derived from the co-occurrences of face tracks. To depict such a constraint, we can define a binary matrix $\mathbf{N} \in \mathbb{R}^{M \times M}$ where $\mathbf{N}_{ab}=1$ if $\mathcal{T}_a$ and $\mathcal{T}_b$ co-occurs and therefore cannot share an identical ID, and $\mathbf{N}_{ab}=0$ for the rest. Consequently, we can describe the cannot-link as a set $\mathcal{N}=\{(\mathbf{x}_i^a, \mathbf{x}_j^b) \mid \forall \mathbf{x}_i^a \in \mathcal{T}_a, \forall \mathbf{x}_j^b \in \mathcal{T}_b, \mathbf{N}_{ab}=1 \}$, and similarly we know that $\mathbbm{1}(\mathbf{x}_i, \mathbf{x}_j)=0, \forall (\mathbf{x}_i, \mathbf{x}_j) \in \mathcal{N}$. 

To start with, most video face clustering works \cite{sharma2019self, Kalogeiton2020bmcv, sharma2020clustering, tapaswi2019video} employ a  certain feature extractor to obtain the facial embedding of each detected facial image, typically through a pre-trained backbone ConvNet. Denote the backbone network as $\op{CNN}$, and denote the embedding of a facial image $\mathbf{x}_i$ as $\mathbf{e}_i \in \mathbb{R}^e$ where $e$ is the embedding dimensionality, we have $\mathbf{e}_i=\op{CNN}(\mathbf{x}_i)$. Note that this backbone network is usually used as it is and is left untouched. 
A Siamese MLP, denoted as $\op{MLP}$, is proposed in \cite{sharma2019self, sharma2020clustering, tapaswi2019video} to project $\mathbf{e}_i$ into a lower-dimensional latent space, i.e. $\mathbf{z}_i=\op{MLP}(\mathbf{e}_i)$ where $\mathbf{z}_i \in \mathbb{R}^z$ and $z<e$,  
and contrastive learning \cite{hadsell2006dimensionality} is used to train the MLP. There are various ways to constitute the positive/negative samples like SSiam \cite{sharma2019self} or CCL \cite{sharma2020clustering}, and we adopt TSiam \cite{sharma2019self} that involves must-links and cannot-links as the example here. An image pair $(\mathbf{x}_i, \mathbf{x}_j)$ is sampled either from the must-link set $\mathcal{M}$ or from the cannot-link set $\mathcal{N}$, with their embedding extracted as $(\mathbf{e}_i, \mathbf{e}_j)$, and the contrastive learning loss $\mathcal{L}_{CT}$ can be formulated as 
\begin{equation}
\label{eq: baseline_contrastive}
\mathcal{L}_{CT} = \frac{y}{2}\norm[\small]{\mathbf{z}_i-\mathbf{z}_j}^2_2 + \frac{(1-y)}{2}\{\max(g - \norm[\small]{\mathbf{z}_i-\mathbf{z}_j}_2, 0)\}^2
\end{equation}
where $\mathbf{z}_i=\op{MLP}(\mathbf{e}_i)$, $\mathbf{z}_j=\op{MLP}(\mathbf{e}_j)$, $y=1$ if $(\mathbf{x}_i, \mathbf{x}_j) \in \mathcal{M}$, $y=0$ if $(\mathbf{x}_i, \mathbf{x}_j) \in \mathcal{N}$, and $g > 0$ is the margin. During the evaluation, the learned $\op{MLP}$ is frame-wisely applied to project the extracted embedding of facial images, and a simple temporal aggregation like average is utilised to summarise the projected embeddings of each face track into a single one, followed by a Hierarchical Agglomerative Clustering (HAC) to gain final clustering predictions.

Note that the contrastive learning in Eq. \ref{eq: baseline_contrastive} is performed at the frame level, and a simple temporal aggregation may ignore the temporally-varying property of the video. To overcome those limitations, we propose to obtain a video-level representation through a video-centralised transformer.

\subsection{System Framework}
The overall framework of the proposed video-centralised transformer is illustrated in Fig. \ref{fig: framework}, where the major improvement comes from the employment of the transformer and the video-centralised learning. Following the common practice in this field, we first extract the embedding for each facial image in a track $\mathcal{T}_a$ using a pre-trained ConvNet $\op{CNN}$. This can written as $\mathcal{E}_a=\{\mathbf{e}_1^a,\mathbf{e}_2^a, ..., \mathbf{e}_n^a\}$ where $\mathcal{E}_a$ represent the track $\mathcal{T}_a$ at embedding level and $\mathbf{e}_i^a=\op{CNN}(\mathbf{x}_i^a)$. 

During the training stage (Fig. \ref{fig: framework} Top), a certain temporal augmentation technique, denoted as $\op{\tau}$, is applied on $\mathcal{E}_a$ to sample a clip $\mathcal{E}_a^{\op{\tau}}$ out of it, i.e. $\mathcal{E}_a^{\op{\tau}}=\op{\tau}(\mathcal{E}_a)$. The sampled clip is then fed into a transformer encoder to learn the video representation in a self-attentive fashion. We generally follow the transformer encoder architecture in ViT \cite{dosovitskiy2020image} which stacks several Multiheaded Self-Attention (MSA) layers together with an MLP head on top, but we also have made some modifications to suit our needs. 
The output of the MLP head is in the new latent space with reduced dimensionality, while the transformer is trained with the proposed video-centralised learning in this space.

The idea of video-centralised learning is based on the simple intuition that for each track $\mathcal{T}_a$, a distinct video centre $\mathbf{c}_a$ is maintained in the latent space. The transformer-based representation of a sampled clip $\mathcal{E}_a^{\op{\tau}}$ should be attracted to $\mathbf{c}_a$, which is an implicit way of utilising the must-link constraints. As for the cannot-link constraints, if two face tracks  $\mathcal{T}_a$ and $\mathcal{T}_b$ co-occur and are known to have exclusive IDs, i.e. $\mathbf{N}_{ab}=1$, we can push the representation of clip $\mathcal{E}_a^{\op{\tau}}$ to be far away from $\mathcal{T}_b$'s video centre $\mathbf{c}_b$, and similarly for $\mathcal{E}_b^{\op{\tau}}$'s representation and $\mathbf{c}_a$. The purpose of video-centralised learning is to enforce the video-level representation from the transformer to be more discriminative, more compact, and most importantly, more \textit{centralised}. 

During the evaluation stage, we discard the sampling step and make use of the whole track for predictions. As shown in Fig. \ref{fig: framework} (Bottom), the facial images are first embedded via the pre-trained ConvNet, which are subsequently input into the transformer without any sampling. 
The output embedding of the transformer encoder without the MLP head is fetched as the video-level representation, and then a HAC is applied to get the final clustering results, following the practices in \cite{sharma2019self, sharma2020clustering}. 

\subsection{Video-level representation with transformer}
The transformer used in our work generally follows the architecture of ViT \cite{dosovitskiy2020image}, yet several modifications have been made to suit our task. In particular, there is no linear projection layer operating on image patches in our transformer, since we have already used a ConvNet to extract embedding. 
We also discard the learnable Positional Embedding (PE) in ViT mainly due to the essence of the studied task.
Unlike other video recognition tasks, the awareness of temporal ordering is not a priority in video face clustering; for instance, if we shuffle the frames in a face track, we still would like to retain the correct ID prediction to be unaffected by this permutation. 
Eliminating PE can simplify the training process and improve efficiency, and we have achieved better performance than state-of-the-art methods without it. 
To verify this assumption, we have performed an ablation study on the effects of PE strategies in Section \ref{sec:ablation_study_pe}.

Formally, given a face track $\mathcal{T}_a$, and let $\mathcal{E}_a$ be the set of its facial embedding extracted by a pre-trained ConvNet $\op{CNN}$, i.e. $\mathcal{E}_a=\{\mathbf{e}_1^a,\mathbf{e}_2^a, ..., \mathbf{e}_n^a\}$ where $\mathbf{e}_i^a=\op{CNN}(\mathbf{x}_i^a)$. We first apply a certain temporal augmentation $\op{\tau}$ to get a clip $\mathcal{E}_a^{\op{\tau}}$. Since the augmentation method is not the focus of this work, we use a simple sampling method $\op{\tau}$, i.e. to sample a clip with consecutive frames at arbitrary length, which can be written as 
\begin{equation}
\label{eq:tau}
\mathcal{E}_a^{\op{\tau}} = \op{\tau}(\mathcal{E}_a)
= \{\mathbf{e}_i^a, \mathbf{e}_{i+1}^a, ..., \mathbf{e}_{i+j}^a\}
\end{equation}
where $i\in\mathbb{Z}_{>0}$ and $j\in\mathbb{Z}_{>0}$ are uniformly sampled from $[1, n)$ and $[1, n-i]$, respectively. 

Similarly to the \verb|[class]| token used by BERT \cite{devlin2018bert} and ViT \cite{dosovitskiy2020image}, we pre-append a learnable embedding token $\mathbf{e}^0$ to $\mathcal{E}_a^{\op{\tau}}$ before feeding it into the transformer encoder, i.e. the input turns into $[\mathbf{e}^0, \mathcal{E}_a^{\op{\tau}}]$, and the state of $\mathbf{e}^0$ at the encoder output side is used as the representation of the track. 

In the transformer, each Multiheaded Self-Attention (MSA) layer is accomplished with certain LayerNorm (LN) layers, MLP layer and Residual Connections. Please refer to \cite{vaswani2017attention, dosovitskiy2020image} for a more detailed illustration. Denote the MSA layer as $\op{MSA}$, the LayerNorm as $\op{LN}$, the MLP layer as $\op{MLP}$, the MLP head at the top as $\op{MLH}$, and assume there are a total of $L$ MSA layers, 
the video-level representation of the sampled clip $\mathcal{E}_a^{\op{\tau}}$, denoted as $\mathbf{z}_a^{\op{\tau}}$, can be explained as
\begin{align}
\mathbf{z}_a^{0} &= [\mathbf{e}^0, \mathcal{E}_a^{\op{\tau}}] = [\mathbf{e}^0, \mathbf{e}_i^a, \mathbf{e}_{i+1}^a, ..., \mathbf{e}_{i+j}^a],  &&  \label{eq: z0} \\
(\mathbf{z}^{\ell}_a)^{\prime} &= \op{MSA}(\op{LN}(\mathbf{z}_a^{\ell-1})) + \mathbf{z}_a^{\ell-1}, && \ell=1 \ldots L  \label{eq: z1} \\
\mathbf{z}_a^{\ell} &=  \op{MLP}(\op{LN}((\mathbf{z}^{\ell}_a)^{\prime}))+(\mathbf{z}^{\ell}_a)^{\prime}, && \ell=1 \ldots L  \label{eq: z2} \\
\mathbf{z}_a^{\op{\tau}} & = \op{MLH}(\op{LN}({\mathbf{z}_a^{L}}[0]))  \label{eq: z3}
\end{align}
where $\mathbf{z}_a^{L}[0]$ denotes the state of the encoder's output that is related with the learnable embedding $\mathbf{e}^0$. 
With video representation $\mathbf{z}_a^{\op{\tau}}$ obtained, we can therefore train the transformer model with the proposed video-centralised learning. Note that we discard the MLP head $\op{MLH}$ during the evaluation stage and $\mathbf{z}_a^{L}[0]$ is fetched as the representation instead. 

For the simplicity of notations, let $\op{f}_{\boldsymbol{\theta}}$ be the transformer model described in Eq. \ref{eq: z0} - \ref{eq: z3} where $\boldsymbol{\theta}$ denotes its learnable parameters, this process can be simplified as $\mathbf{z}_a^{\op{\tau}}=\op{f}_{\boldsymbol{\theta}} (\mathcal{E}_a^{\op{\tau}})$.

\subsection{Video-centralised learning}
In this work, we present the video-centralised learning based on the hinge-loss-like contrastive learning  \cite{hadsell2006dimensionality} (described in Eq. \ref{eq: baseline_contrastive}) which is widely adopted by prior works \cite{sharma2019self, sharma2020clustering}. However, it is worth noticing that the idea of introducing video centres can also be potentially applied to InfoNCE \cite{oord2018representation}, a cross-entropy-based loss, with appropriate modifications.

Let $\mathbf{c}_a$ be the video centre that is assigned with face track $\mathcal{T}_a$ in the target latent space, we aim to attract the representations of any sampled clip $\mathcal{E}_a^{\op{\tau}}$ to $\mathbf{c}_a$. This can be written as 
\begin{equation}
\label{eq: attract}
\mathcal{L}_{att} = \norm[\small]{\mathbf{z}_a^{\op{\tau}}-\mathbf{c}_a}_2
\end{equation}
where $\mathcal{L}_{att}$ denotes the loss from this attraction process.
This is essentially the utilisation of the must-links at the video level. Note that there is no need to perform the pairwise sampling as in Eq. \ref{eq: baseline_contrastive}, as long as $\mathbf{c}_a$ can be established.
As for the cannot-links, let $\mathcal{T}_b$ be a co-occurring face track with $\mathcal{T}_a$, i.e. $\mathbf{N}_{ab}=1$, and let $\mathbf{c}_b$ be the video centre of $\mathcal{T}_b$. We require the  representation of $\mathcal{T}_a$ to be repelled from $\mathbf{c}_b$ (and similarly for $\mathcal{T}_b$ and $\mathbf{c}_a$), which can be described as
\begin{equation}
\label{eq: repel}
\mathcal{L}_{rpl} = \max(g-\norm[\small]{\mathbf{z}_a^{\op{\tau}}-\mathbf{c}_b}_2,0) 
\end{equation}
where $\mathcal{L}_{rpl}$ is the loss function for this repelling process, and $g>0$ is the margin. The definition of $\mathcal{L}_{rpl}$ also eliminates the pre-requisite of pairwise sampling in Eq. \ref{eq: baseline_contrastive}. 

Putting Eq. \ref{eq: attract} and Eq. \ref{eq: repel} together, the video-centralised loss $\mathcal{L}$ can be written as 
\begin{align}
\mathcal{L} &=\frac{y}{2}\mathcal{L}_{att}+\frac{(1-y)}{2}\mathcal{L}_{rpl},  \nonumber \\
 &= \frac{y}{2}\norm[\small]{\mathbf{z}_a^{\op{\tau}}-\mathbf{c}_a}_2 + \frac{(1-y)}{2} \max(g-\norm[\small]{\mathbf{z}_a^{\op{\tau}}-\mathbf{c}_b}_2,0) \label{eq: VC loss}
\end{align}
where $y=1$ if $\mathbf{z}_a$ is to be attracted to its own centre, and $y=0$ if $\mathbf{z}_a^{\op{\tau}}$ is to be repelled from a negative centre. In theory, $\mathcal{L}$ can also have a triplet form, i.e. $\mathbf{z}_a^{\op{\tau}}$ being attracted and repelled at the time, but we stick to Eq. \ref{eq: VC loss} in this work for the ease of explanations.

How to represent the video centre $\mathbf{c}_a$ is a core issue here. An intuitive way is that we can perform a forward propagation on the whole embedded track $\mathcal{E}_a$ and use the output as the centre, i.e. 
\begin{align}
\label{eq: whole_track_center}
\mathbf{c}_a = \op{f}_{\boldsymbol{\theta}} (\mathcal{E}_a).
\end{align}
This centre representation covers the whole face track and therefore can be more unbiased than using representations of sampled clips. In this work, Eq. \ref{eq: whole_track_center} is mainly used to initialise or to re-establish the video centres for each face track during the training stage. However, it is computationally infeasible to apply Eq. \ref{eq: whole_track_center} on a per-training-step basis, since each call of Eq. \ref{eq: whole_track_center} will require a forward pass of transformers on all face tracks, which will lead to extremely high time complexity considering the number of potential steps.

Therefore, we employ a more computationally-efficient way to represent video centres and to optimise the video-centralised loss in Eq. \ref{eq: VC loss}. A two-step optimisation process is involved to update 
the transformer parameters $\boldsymbol{\theta}$ and the video centre representations in an iterative manner following \cite{yang2017towards}, after those video centres being initialised with Eq. \ref{eq: whole_track_center}. 

\subsection{Optimisation} 
Following a full forward propagation, the first optimisation step is to minimise $\mathcal{L}$ with respect to (w.r.t.) $\mathcal{E}_a^{\op{\tau}}$ through updating the transformer's parameters $\boldsymbol{\theta}$, while the video centres $\mathbf{c}_a$ and $\mathbf{c}_b$ are seen as constant values. This is achieved through an SGD-based back-propagation and can be depicted as 
\begin{align}
\label{eq: update_theta}
\boldsymbol{\theta} \leftarrow \boldsymbol{\theta} - \xi \nabla_{\boldsymbol{\theta}}{\mathcal{L}} 
\end{align}
where $\xi$ is the learning rate, and  $\nabla_{\boldsymbol{\theta}}{\mathcal{L}}=\pdv{\mathcal{L}}{\boldsymbol{\theta}}=\pdv{\mathcal{L}}{\mathbf{z}_a^{\op{\tau}}}\pdv{\mathbf{z}_a^{\op{\tau}}}{\boldsymbol{\theta}}$. With $\mathbf{c}_a$ and $\mathbf{c}_b$ frozen, we can easily compute the partial derivative component on the left-hand side as 
\begin{align}
\pdv{\mathcal{L}}{\mathbf{z}_a^{\op{\tau}}} = \frac{y}{2}\frac{(\mathbf{z}_a^{\op{\tau}}-\mathbf{c}_a)}{\|\mathbf{z}_a^{\op{\tau}}-\mathbf{c}_a\|_2}- \frac{(1-y)}{2}\frac{\mathbbm{1}(\mathcal{L}_{rpl}>0)(\mathbf{z}_a^{\op{\tau}}-\mathbf{c}_b)}{\|\mathbf{z}_a^{\op{\tau}}-\mathbf{c}_b\|_2} \nonumber
\end{align}
where $\mathbbm{1}(\mathcal{L}_{rpl}>0)=1$ if $\mathcal{L}_{rpl}>0$ and $\mathbbm{1}(\mathcal{L}_{rpl}>0)=0$ otherwise.
Another component $\pdv{\mathbf{z}_a^{\op{\tau}}}{\boldsymbol{\theta}}$ can be obtained with chain rules and is omitted here.  

The second step is to update the video centres with $\boldsymbol{\theta}$ frozen. To save computations, we use the pre-computed values of $\mathbf{z}_a^{\op{\tau}}$ from the latest forward propagation to perform this update. Particularly, we aim to minimise $\mathcal{L}$ w.r.t. $\mathbf{c}_a$, $\mathbf{c}_b$ through a SGD-like algorithm, considering $\mathbf{z}_a^{\op{\tau}}$ from the last forward propagation as constant values. The optimisation of $\mathbf{c}_a$ can be written as 
\begin{align}
\mathbf{c}_a &\leftarrow \mathbf{c}_a - \eta \nabla_{\mathbf{c}_a}{\mathcal{L}},  \label{eq: update_cen_a} \\
\nabla_{\mathbf{c}_a}{\mathcal{L}} &=  \pdv{\mathcal{L}}{\mathbf{c}_a}
=\frac{y}{2} 
\frac{(\mathbf{c}_{a}-\mathbf{z}_a^{\op{\tau}})}
{\norm[\small]{\mathbf{z}_a^{\op{\tau}}-\mathbf{c}_{a}}_2}  \nonumber
\end{align}
where $\eta$ is the learning rate.  
Similarly, the update of $\mathbf{c}_b$ can be described as  
\begin{align}
\mathbf{c}_b &\leftarrow \mathbf{c}_b - \eta \nabla_{\mathbf{c}_b}{\mathcal{L}}, \label{eq: update_cen_b} \\
\nabla_{\mathbf{c}_b}{\mathcal{L}} &=  \pdv{\mathcal{L}}{\mathbf{c}_b}
=\frac{(1-y)}{2}\mathbbm{1}(\mathcal{L}_{rpl}>0) \frac{(\mathbf{z}_a^{\op{\tau}}-\mathbf{c}_{b} )}{\norm[\small]{\mathbf{z}_a^{\op{\tau}}-\mathbf{c}_{b}}_2}. \nonumber
\end{align}

The updating of transformer parameters (Eq. \ref{eq: update_theta}) and video centres (Eq. \ref{eq: update_cen_a} and \ref{eq: update_cen_b}) are performed iteratively during each training step. Since we use the pre-computed $\mathbf{z}_a^{\op{\tau}}$ to update video centres in Eq. \ref{eq: update_cen_a} and \ref{eq: update_cen_b}, the computations are significantly accelerated when compared with Eq. \ref{eq: whole_track_center}. In practice, in addition to the SGD-like updates in Eq. \ref{eq: update_cen_a} and \ref{eq: update_cen_b}, 
we also invoke Eq. \ref{eq: whole_track_center} to
regularly re-compute video centres after certain steps to obtain a more unbiased centre representation. The algorithm to optimise the proposed video-centralised learning is shown in Algorithm \ref{alg:optimise}.

\begin{algorithm}[t!]
\caption{Optimisation of video-centralised learning}
\label{alg:optimise}
\begin{algorithmic}[1]
\Require Training steps $T_1$, an interval $T_2 < T_1$
\State Compute video centres with Eq. \ref{eq: whole_track_center}  \label{algo_step: init}
\For{$t=1:T_1$}
\State Update network parameters $\boldsymbol{\theta}$ using Eq. \ref{eq: update_theta} 
\State Update video centres using Eq. \ref{eq: update_cen_a} and \ref{eq: update_cen_b}  
\If{$t \bmod T_2=0$}  \label{algo_step: re-compute-1}
\State Repeat step \ref{algo_step: init}  \label{algo_step: re-compute-2}
\EndIf 
\EndFor
\end{algorithmic}
\end{algorithm}

\section{Experiments}
\subsection{Dataset}
Our experiments are conducted on two video clustering datasets, the Big Bang Theory (BBT) dataset \cite{roy2014tvd, tapaswi2019video} consisting of six episodes from the first season of the TV show \say{Big Bang Theory}, and our EasyCom-Clustering with 22 egocentric videos recorded in a simulated restaurant setting. 

BBT dataset is initially released by \cite{roy2014tvd, tapaswi2012knock} with annotations for six characters from the first-season TV shows \textit{Big Bang Theory}. This dataset is further enhanced in \cite{sharma2019self, tapaswi2019video} by annotating more characters. 
In this work, we use the most recent version of BBT \cite{tapaswi2019video}
There are a total of 6 episodes in this dataset, indexed from \say{s01e01} to \say{s01e06} where \say{s} represents the season and \say{e} denotes the episode. Each BBT episode lasts for around $20$ minutes, with a total of $3\,908$ annotated face tracks of $232\,176$ facial images detected from $103$ characters.
The facial embeddings are extracted with a VGGFace2 \cite{cao2018vggface2} model of SE-ResNet-50 architecture \cite{hu2018squeeze} that is pre-trained on MS1M-V0 \cite{guo2016ms} and then is fined-tuned on VGGFace2 \cite{cao2018vggface2} using soft-max loss. 
In the BBT dataset, episodes are not fully subject-independent of each other, and some characters shown in one episode may still be found in another episode.
The average duration of each track is $59$ frames, which is approximately $2$ seconds for videos at $30$ FPS. We evaluate the performance of various methods on each of the six episodes for completeness. 

\begin{figure}[b]
\centering
\includegraphics[width=0.99\linewidth]{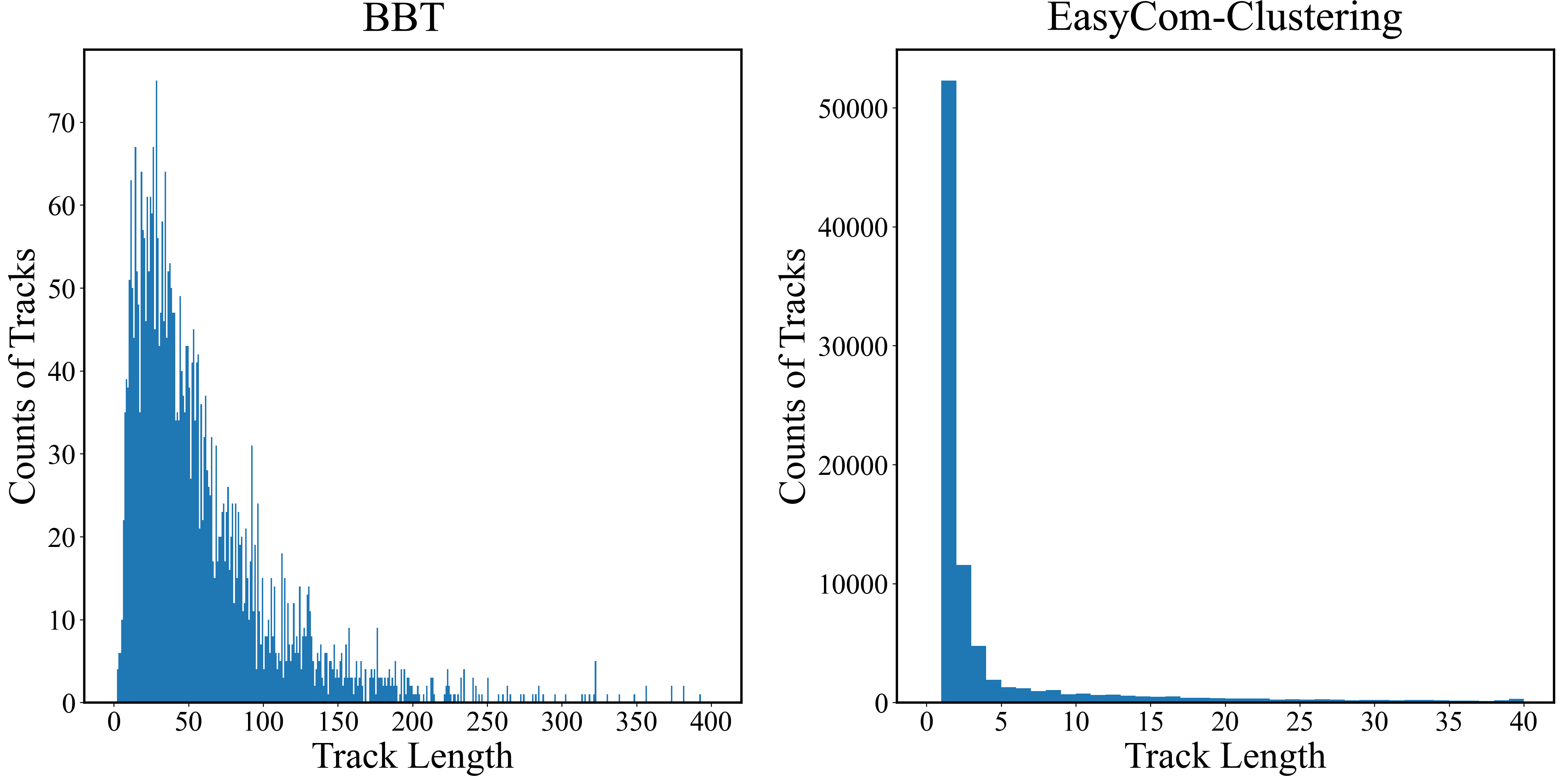} 
\caption{The histogram plot of face track lengths in BBT (\textit{Left}) and EasyCom-Clustering (\textit{Right}).}
\label{fig: len_hist}
\end{figure}

We also construct the EasyCom-Clustering dataset as another dataset for evaluation. This dataset contains $22$ sessions of egocentric recordings collected in a simulated restaurant setting, and each session lasts for around $30$ minutes. $15$ out of the $22$ sessions are already used in the recently released EasyCom \cite{donley2021easycom} dataset on multi-modal audio-visual analysis under noisy environments. In this work, we make use of all the $22$ sessions, and we have detected, embedded and annotated a total of $94\,047$ face tracks with $1\,623\,633$ facial images from $53$ participants, 
using RetinaFace \cite{deng2020retinaface} as the face detector and an Arcface \cite{deng2019arcface} model for facial embedding. The architecture of this ArcFace model is iResNet-100 \cite{deng2019arcface}, and it is pre-trained on the MS1M-ArcFace dataset \cite{deng2019arcface} using the ArcFace loss \cite{deng2019arcface}.
Unlike the noisy multi-modal analysis problem in EasyCom, this dataset is exclusively designed for identity-related problems. Besides, EasyCom-Clustering is mostly subject-independent between sessions, i.e. except for one subject that appears in every session, all other characters can only be found in one of the $22$ sessions.

EasyCom-Clustering has significantly different track length distributions when compared with BBT dataset, as shown in Fig. \ref{fig: len_hist}. The majority duration of BBT face tracks falls between $10$ and $150$, while most face tracks in EasyCom-Clustering last for less than $20$ frames. The average duration of the face tracks in EasyCom-Clustering is $17$, significantly less than the $59$ of BBT. This is not surprising, considering that the gaze points from the egocentric view can change much more frequently than a third-person camera, resulting in shorter face tracks. Similar to BBT, we also evaluate our method and different methods on each of the $22$ sessions of EasyCom-Clustering, and those sessions are indexed from \say{V01} to \say{V22} in this paper.
 
\subsection{Experimental Setup}
\subsubsection{Evaluation Metrics}
In this work, we examine the quality of the video representation under two different clustering settings, i.e. the total cluster number is either 1). known, or 2). unknown. This is achieved by applying different stopping criteria to HAC. For known cluster numbers, the agglomerative merging will be stopped when this cluster number is achieved, while for the case of unknown cluster numbers, a pre-defined distance threshold is required as the stopping criterion. 

We, therefore, adopt slightly different metrics for those two cases. Currently, there are two commonly used metrics in previous works, the Weighted Clustering Purity (WCP) \cite{tapaswi2014total, zhang2016joint} and the Normalised Mutual Information (NMI) \cite{schutze2008introduction}. WCP evaluates the purity of a predicted cluster by assigning it the class of its most-frequent samples. As discussed in previous works \cite{sharma2019self, Kalogeiton2020bmcv}, WCP is a fair evaluation metric if the cluster number is pre-known and all methods predict the same number of clusters, therefore we also adopt this metric for the known-cluster-number experiments. In addition to WCP, we also employ NMI in this setting. Given a cluster prediction $\mathbf{P}$ and the ground-truth $\mathbf{Y}$, the NMI can be computed as $\textnormal{NMI}=\frac{2\op{I}(\mathbf{Y},\mathbf{P})}{\op{H}(\mathbf{P})+\op{H}(\mathbf{Y})}$ where $\op{I}$ refers to the mutual information and $\op{H}$ is the entropy. 

For unknown cluster experiments, situations are somehow different. The cluster number is no long pre-known, and therefore different algorithms can predict a different number of clusters. The one closest to the true cluster number should be considered as the most desirable. Therefore, we consider the absolute differences between the predicted and true cluster numbers as an important metric, abbreviated as \say{\#C DIF}. As shown in related works \cite{Kalogeiton2020bmcv, tapaswi2019video}, WCP is a biased metric for unknown clustering tasks, e.g. the WCP value will be $1.0$ if each sample is assigned with a separate cluster number, and therefore it is not a suitable metric for this experimental setting. On the other hand, NMI is still a fair metric that can better reveal the trade-offs between cluster number and clustering accuracy. 
Therefore we incorporate NMI in addition to the absolute differences of cluster number (\say{\#C DIF}) to evaluate the clustering performance with unknown cluster numbers.

\subsubsection{Baselines}
We mainly compare the performance of our video-centralised transformer (VC TRSF.) with the following five baselines: 

\begin{enumerate}
\item The simple video face clustering framework in \cite{Sharma2017ASimple}, abbreviated as \say{Temp. AVG}, in which the extracted facial embedding of a face track is directly averaged along the temporal dimension without any learning models like MLP.

\item SSiam \cite{sharma2019self} that trains a Siamese MLP as in Eq. \ref{eq: baseline_contrastive}, while the positive/negative samples are generated through the mining of the challenging data.

\item CCL \cite{sharma2020clustering} which also trains a Siamese MLP with Eq. \ref{eq: baseline_contrastive}, while the clustering results from FINCH \cite{sarfraz2019efficient} are used as the pseudo labels.

\item TSiam \cite{sharma2019self} that learns the Siamese MLP with Eq. \ref{eq: baseline_contrastive} via exploiting the must-links $\mathcal{M}$ and cannot links $\mathcal{N}$.

\item A transformer that is trained using the naive extension framework of contrastive learning as illustrated in Fig. 1 (\textit{Middle}). The training setting is mostly identical to our method, except that it is learned in a pairwise manner similar to Eq. \ref{eq: baseline_contrastive} without using video centres. This baseline is abbreviated as \say{CT TRSF.} for convenience. 
\end{enumerate}

\begin{table*}[ht!]
\caption{The ablation study on video \protect\say{V04} of EasyCom-Clustering with known cluster number. The baseline model in the first row is a transformer trained with the paradigm in Fig. \ref{fig: idea_explain} (\textit{Middle}), i.e. a CT TRSF.
The option \protect\say{a} in each component is considered as the naive solution and are gradually replaced by others to illustrate the improvement.}
\centering
\scriptsize
\begin{tabularx}{0.86\linewidth}{viiiyiiiii}
\toprule
  & \multicolumn{2}{c}{\textbf{Temporal Aug.} $\op{\tau}$} & \multicolumn{3}{c}{\textbf{Centre Representation}} & \multicolumn{2}{c}{\textbf{Centre Learning Rate $\eta$}} &
  \multicolumn{2}{c}{\textbf{EasyCom-Clustering}} \\
 & \multicolumn{1}{c}{a} & \multicolumn{1}{c}{b} & \multicolumn{1}{c}{a} & \multicolumn{1}{c}{b} & \multicolumn{1}{c}{c} & \multicolumn{1}{c}{a} & \multicolumn{1}{c}{b} & \multicolumn{2}{c}{V04} \\
\multirow{-3}{*}{\textbf{\shortstack{Video \\Centres}}} & Uniform & Consect. & Last Forw. & Alg. \ref{alg:optimise} Incomp. & Alg. \ref{alg:optimise} & Incremental & Proportional & NMI (\%) & WCP (\%) \\
\cmidrule(l{4pt}r{0pt}){1-1} 
\cmidrule(l{10pt}r{10pt}){2-3} 
\cmidrule(l{7pt}r{12pt}){4-6}
\cmidrule(l{8pt}r{0pt}){7-8}
\cmidrule(l{8pt}r{8pt}){9-10} 
- & - & - & - & - & - & - & - & 85.20 & 95.19 \\
\checkmark & \checkmark & - & \checkmark & - & - & \checkmark & - & 83.54 & 93.89 \\
\checkmark & - & \checkmark & \checkmark & - & - & \checkmark & - & 83.97 & 94.31 \\
\checkmark & - & \checkmark & - & \checkmark & - & \checkmark & - & 86.20 & 95.94 \\
\checkmark & - & \checkmark & - & - & \checkmark & \checkmark & - & 87.19 & 96.30 \\
\checkmark & - & \checkmark & - & - & \checkmark & - & \checkmark & \textbf{90.69} & \textbf{97.83} \\
\bottomrule
\end{tabularx}
\label{tab: ablation} 
\end{table*}

\subsubsection{Training Settings}
Following the optimisation process in Algorithm \ref{alg:optimise}, there are essentially two sets of optimisation settings in our system, one for optimising the transformer parameters $\boldsymbol{\theta}$ and another for updating video centres. For the former, we utilise SGD with momentum as the optimiser with a weight decay of $1\mathrm{e}{-5}$,
and leverage the OneCycle learning rate policy \cite{smith2019super} to implement $\xi$. As for the latter, we directly use Eq. \ref{eq: update_cen_a} and \ref{eq: update_cen_b} to update the video centres, and its learning rate is defined as $\eta=p\xi$ where $p > 0$ is a constant number. The intuition is that we should keep the step size of updating video centres to be proportional to that of updating network parameters. There are other ways of designing the video learning rate $\eta$, but we find this way can work best in practice. 

We train a total of $900$ epochs in each training session, with the first $400$ epochs as the warm-up stage. The maximum learning rate in the OneCycle scheduler is set to $5.1\mathrm{e}{-4}$.
A batch size of $512$ is used, and a maximum length of $90$ is posed on the sampled clip to ensure reasonable memory usage.
For each training epoch, we sampled $10$ clips out of each face track to be attracted to its own centre, while $16$ clips are sampled per track to be repelled to one of its negative video centres. The margin value $g$ in Eq. \ref{eq: VC loss} is empirically set to be $1.0$. Besides, we use Eq. \ref{eq: whole_track_center} to initialise the video centres at the beginning of each training session, while for every $50$ epochs we re-compute video centres with Eq. \ref{eq: whole_track_center}.

We have employed a transformer architecture consisting of $4$ MSA layers, and the number of heads in each MSA layer is $16$. The input embedding dimension varies between datasets. A $256$-dimensional embedding is used on BBT, while the input dimensionality on EasyCom-Clustering is $512$. The MLP head consists of a LayerNorm and a Fully-Connected Layer to reduce the dimensionality to $2$. All transformer models are trained from scratch without using any pre-trained checkpoints. Since we are working on a self-supervised problem, we use the same video for both training and evaluation. For validation on the BBT dataset, we leverage S-Dbw \cite{halkidi2001clustering} computed on the same video as training/evaluation. S-Dbw is a kind of internal clustering index \cite{liu2010understanding} that can be computed without true clustering labels, and we choose the checkpoint with the best S-Dbw value. As for the EasyCom-Clustering dataset, we simply choose the final checkpoint of each training session for evaluation. For unknown cluster experiments, the thresholds to stop the agglomerative merging for different methods are selected on a per-method and per-dataset basis and are determined by fine-tuning on a certain video of each dataset, i.e. video \say{s01e02} on BBT dataset, and video \say{V09} on EasyCom-Clustering.

\begin{table}[ht!]
\caption{The ablation study of different Positional Embedding (PE) options on \protect\say{V08} of EasyCom-Clustering dataset (known cluster).}
\centering
\scriptsize
\begin{tabularx}{0.98\linewidth}{kkjj}
\toprule
\multirow{2}{*}{\textbf{Method}} & \multirow{2}{*}{\shortstack{\textbf{Positional} \\ \textbf{Embedding}}} & \multicolumn{2}{c}{V08} \\
 &  & NMI (\%) & WCP (\%) \\
\cmidrule(lr){1-1} \cmidrule(lr){2-2} \cmidrule(lr){3-4}  
Temp. AVG \cite{Sharma2017ASimple} & \multirow{5}{*}{-} & 61.89 & 70.88 \\
SSiam \cite{sharma2019self} &  & 74.79 & 88.99 \\
CCL \cite{sharma2020clustering} &  & 83.99 & 93.98 \\
TSiam \cite{sharma2019self} &  & 82.17 & 92.61 \\
CT TRSF. &  & 82.57 & 94.05 \\
\cmidrule(lr){1-1} \cmidrule(lr){2-2} \cmidrule(lr){3-4}
\multirow{3}{*}{VC TRSF. (Ours)} & No PE & \textbf{86.70} & 95.10 \\
 & Scaled PE \cite{li2019neural} & 84.53 & 95.28 \\
 & Interpolated PE \cite{bertasius2021space} & 85.89 & \textbf{95.33}  \\
\bottomrule
\end{tabularx}
\label{tab: pe_ablation} 
\end{table}

\begin{table*}[ht!]
\caption{Video clustering performance of different methods on BBT dataset with known cluster numbers. The true cluster number of each video is post-appended to the video name.  }
\centering
\scriptsize
\begin{tabularx}{0.8\linewidth}{siiiiiiiiiiii}
\toprule
& \multicolumn{12}{c}{\textbf{BBT} (\%), Known Cluster Number} \\
\cmidrule(l{5pt}r{3pt}){2-13} 
\multirow{2}{*}{\textbf{Method}} & \multicolumn{2}{c}{s01e01 (8)} & \multicolumn{2}{c}{s01e02 (6)} & \multicolumn{2}{c}{s01e03 (26)} & \multicolumn{2}{c}{s01e04 (28)} & \multicolumn{2}{c}{s01e05 (25)} & \multicolumn{2}{c}{s01e06 (37)} \\
 & NMI & WCP & NMI & WCP & NMI & WCP & NMI & WCP & NMI & WCP & NMI & WCP \\
   \cmidrule(l{16pt}r{16pt}){1-1} 
  \cmidrule(l{5pt}r{3pt}){2-3} 
  \cmidrule(l{5pt}r{3pt}){4-5}
  \cmidrule(l{5pt}r{3pt}){6-7}
  \cmidrule(l{5pt}r{3pt}){8-9}
  \cmidrule(l{5pt}r{3pt}){10-11}
  \cmidrule(l{5pt}r{3pt}){12-13} 
Temp. AVG \cite{Sharma2017ASimple} & 95.74 & 97.87 & 87.80 & 95.45 & 89.46 & 96.82 & 83.74 & 95.60 & 91.57 & 97.52 & 83.10 & 91.79 \\
SSiam \cite{sharma2019self} & 97.49 & 99.39 & 93.44 & 98.54 & 95.92 & 97.12 & 88.01 & 96.25 & 94.35 & 97.90 & 84.87 & 92.86 \\
CCL \cite{sharma2020clustering} & 98.06 & 98.63 & 96.27 & 99.51 & 95.30 & 97.27 & 91.03 & \textbf{97.06} & 96.78 & 97.71 & \textbf{86.45} & 92.50 \\
TSiam \cite{sharma2019self} & 97.06 & 98.48 & 95.46 & 99.51 & \textbf{96.91} & 97.12 & \textbf{91.55} & 96.08 & \textbf{97.40} & 98.28 & 86.30 & 92.74 \\
CT TRSF. & 97.51 & 99.09 & 97.32 & 99.19 & 93.01 & 96.52 & 86.70 & 95.92 & 92.63 & 98.09 & 84.02 & 91.79 \\
  \cmidrule(l{5pt}r{3pt}){2-3} 
  \cmidrule(l{5pt}r{3pt}){4-5}
  \cmidrule(l{5pt}r{3pt}){6-7}
  \cmidrule(l{5pt}r{3pt}){8-9}
  \cmidrule(l{5pt}r{3pt}){10-11}
  \cmidrule(l{5pt}r{3pt}){12-13} 
VC TRSF. (Ours) & \textbf{98.52} & \textbf{99.39} & \textbf{99.32} & \textbf{99.84} & 96.60 & \textbf{97.58} & 91.13 & 96.41 & 94.70 & \textbf{98.47} & 84.91 & \textbf{93.33} \\
\bottomrule
\end{tabularx}
\label{tab: bbt_known} 
\end{table*}
\begin{table*}[ht]
\caption{Video clustering performance of different methods on EasyCom-Clustering dataset with known cluster numbers. The true cluster number of each video is post-appended to the video name. }
\centering
\scriptsize
\begin{tabularx}{0.99\linewidth}{siiiiiiiiiiiiiiiiiiiiii}
\toprule
 & \multicolumn{22}{c}{\textbf{EasyCom-Clustering} (\%), Known Cluster Number} \\
 \cmidrule(l{5pt}){2-23} 
 & \multicolumn{2}{c}{V01 (4)} & \multicolumn{2}{c}{V02 (3)} & \multicolumn{2}{c}{V03 (3)} & \multicolumn{2}{c}{V04 (3)} & \multicolumn{2}{c}{V05 (4)} & \multicolumn{2}{c}{V06 (3)} & \multicolumn{2}{c}{V07 (3)} & \multicolumn{2}{c}{V08 (4)} & \multicolumn{2}{c}{V09 (3)} & \multicolumn{2}{c}{V10 (3)} & \multicolumn{2}{c}{V11 (3)} \\
\multirow{-2}{*}{\textbf{Methods}} & NMI & WCP & NMI & WCP & NMI & WCP & NMI & WCP & NMI & WCP & NMI & WCP & NMI & WCP & NMI & WCP & NMI & WCP & NMI & WCP & NMI & WCP \\
\cmidrule(l{3pt}r{3pt}){1-1} 
 \cmidrule(l{7pt}r{0pt}){2-3}  \cmidrule(l{7pt}r{0pt}){4-5}  \cmidrule(l{7pt}r{0pt}){6-7}
  \cmidrule(l{7pt}r{0pt}){8-9}
   \cmidrule(l{7pt}r{0pt}){10-11}
    \cmidrule(l{7pt}r{0pt}){12-13}
    \cmidrule(l{7pt}r{0pt}){14-15}
    \cmidrule(l{7pt}r{0pt}){16-17}
    \cmidrule(l{7pt}r{0pt}){18-19}
    \cmidrule(l{7pt}r{0pt}){20-21}
    \cmidrule(l{7pt}r{0pt}){22-23}
Temp. AVG \cite{Sharma2017ASimple} & 85.96 & 91.44 & 45.99 & 58.37 & 80.36 & 94.00 & 75.31 & 92.34 & 53.63 & 53.63 & 37.56 & 54.10 & 81.04 & 94.80 & 61.89 & 70.88 & 69.88 & 91.63 & 61.36 & 87.27 & 55.83 & 79.48 \\
SSiam \cite{sharma2019self} & 90.93 & 96.96 & 79.06 & 94.18 & 91.19 & 98.26 & 79.76 & 94.49 & 80.94 & 92.84 & 73.66 & 90.01 & 91.07 & 98.26 & 74.79 & 88.99 & 79.31 & 92.52 & 82.87 & 94.73 & 81.89 & 95.07 \\
CCL \cite{sharma2020clustering} & 94.39 & 98.45 & 80.11 & 94.47 & 92.79 & 98.52 & 83.93 & 94.36 & 89.49 & \textbf{97.41} & 76.73 & 92.13 & 92.07 & 98.39 & 83.99 & 93.98 & 89.38 & 97.07 & 75.67 & 93.20 & 79.47 & 94.18 \\
TSiam \cite{sharma2019self} & 95.25 & 98.79 & 79.69 & 93.53 & 87.65 & 96.60 & 84.44 & 95.03 & \textbf{90.66} & 97.01 & 78.29 & 91.96 & 90.79 & 98.09 & 82.17 & 92.61 & 74.80 & 91.69 & 83.17 & 94.05 & 84.62 & 95.35 \\
CT TRSF. & 90.16 & 95.33 & 74.41 & 91.87 & 94.40 & 98.88 & 85.20 & 95.19 & 80.09 & 91.10 & 76.46 & 92.21 & 91.54 & 98.05 & 82.57 & 94.05 & 85.99 & 94.79 & 81.29 & 91.89 & 80.37 & 93.82 \\
\cmidrule(l{7pt}r{0pt}){2-3}  \cmidrule(l{7pt}r{0pt}){4-5}  \cmidrule(l{7pt}r{0pt}){6-7}
  \cmidrule(l{7pt}r{0pt}){8-9}
   \cmidrule(l{7pt}r{0pt}){10-11}
    \cmidrule(l{7pt}r{0pt}){12-13}
    \cmidrule(l{7pt}r{0pt}){14-15}
    \cmidrule(l{7pt}r{0pt}){16-17}
    \cmidrule(l{7pt}r{0pt}){18-19}
    \cmidrule(l{7pt}r{0pt}){20-21}
    \cmidrule(l{7pt}r{0pt}){22-23}
VC TRSF. (Ours) & \textbf{96.26} & \textbf{99.02} & \textbf{82.25} & \textbf{95.35} & \textbf{94.86} & \textbf{99.05} & \textbf{90.69} & \textbf{97.83} & 87.39 & 96.01 & \textbf{79.81} & \textbf{92.43} & \textbf{93.29} & \textbf{98.59} & \textbf{86.70} & \textbf{95.10} & \textbf{98.17} & \textbf{99.75} & \textbf{89.47} & \textbf{97.52} & \textbf{87.12} & \textbf{95.70} \\
 \midrule \midrule  
 & \multicolumn{2}{c}{V12 (3)} & \multicolumn{2}{c}{V13 (3)} & \multicolumn{2}{c}{V14 (5)} & \multicolumn{2}{c}{V15 (3)} & \multicolumn{2}{c}{V16 (3)} & \multicolumn{2}{c}{V17 (3)} & \multicolumn{2}{c}{V18 (3)} & \multicolumn{2}{c}{V19 (3)} & \multicolumn{2}{c}{V20 (3)} & \multicolumn{2}{c}{V21 (5)} & \multicolumn{2}{c}{V22 (4)} \\
& NMI & WCP  & NMI & WCP & NMI & WCP & NMI & WCP & NMI & WCP & NMI & WCP & NMI & WCP & NMI & WCP & NMI & WCP & NMI & WCP & NMI & WCP \\
 \cmidrule(l{7pt}r{0pt}){2-3}  \cmidrule(l{7pt}r{0pt}){4-5}  \cmidrule(l{7pt}r{0pt}){6-7}
  \cmidrule(l{7pt}r{0pt}){8-9}
   \cmidrule(l{7pt}r{0pt}){10-11}
    \cmidrule(l{7pt}r{0pt}){12-13}
    \cmidrule(l{7pt}r{0pt}){14-15}
    \cmidrule(l{7pt}r{0pt}){16-17}
    \cmidrule(l{7pt}r{0pt}){18-19}
    \cmidrule(l{7pt}r{0pt}){20-21}
    \cmidrule(l{7pt}r{0pt}){22-23}
Temp. AVG \cite{Sharma2017ASimple} & 66.59 & 88.63 & 39.30 & 71.07 & 82.68 & 85.43 & 38.24 & 58.15 & 91.32 & 98.11 & 62.60 & 61.86 & 82.40 & 93.91 & 59.95 & 64.59 & 77.97 & 92.92 & 60.23 & 66.50 & 60.90 & 76.61 \\
SSiam \cite{sharma2019self} & 83.87 & 95.30 & 88.85 & 97.30 & 84.11 & 84.65 & 79.52 & 94.31 & 92.41 & 98.16 & 62.06 & 61.78 & 92.03 & 98.45 & 94.06 & 98.75 & 88.29 & 96.75 & 82.04 & 91.67 & 82.59 & 94.09 \\
CCL \cite{sharma2020clustering} & 86.63 & 96.84 & 88.71 & 97.03 & 89.25 & \textbf{96.25} & 86.85 & 96.94 & 95.45 & 99.10 & 61.12 & 61.86 & 89.38 & 97.45 & 93.86 & 98.66 & 89.46 & 97.80 & 85.36 & 93.49 & 84.44 & 94.01 \\
TSiam \cite{sharma2019self} & 86.85 & 96.56 & 85.36 & 95.67 & 89.34 & 95.59 & 89.49 & 97.32 & 94.84 & 98.90 & 91.00 & 97.72 & 90.87 & 97.81 & 97.62 & 99.58 & 89.61 & 97.70 & \textbf{85.88} & 93.07 & 83.82 & 93.77 \\
CT TRSF. & 81.08 & 93.62 & 87.76 & 96.76 & 89.89 & \textbf{96.25} & 83.99 & 95.23 & 83.60 & 95.03 & 73.13 & 85.65 & 90.68 & 97.90 & 93.41 & 98.43 & 90.97 & 98.07 & 80.42 & 91.65 & 81.20 & 93.07 \\
 \cmidrule(l{7pt}r{0pt}){2-3}  \cmidrule(l{7pt}r{0pt}){4-5}  \cmidrule(l{7pt}r{0pt}){6-7}
  \cmidrule(l{7pt}r{0pt}){8-9}
   \cmidrule(l{7pt}r{0pt}){10-11}
    \cmidrule(l{7pt}r{0pt}){12-13}
    \cmidrule(l{7pt}r{0pt}){14-15}
    \cmidrule(l{7pt}r{0pt}){16-17}
    \cmidrule(l{7pt}r{0pt}){18-19}
    \cmidrule(l{7pt}r{0pt}){20-21}
    \cmidrule(l{7pt}r{0pt}){22-23}
VC TRSF. (Ours) & \textbf{90.89} & \textbf{98.30} & \textbf{92.99} & \textbf{98.61} & \textbf{90.20} & 95.87 & \textbf{92.96} & \textbf{98.55} & \textbf{97.22} & \textbf{99.46} & \textbf{95.37} & \textbf{99.11} & \textbf{93.15} & \textbf{98.63} & \textbf{98.41} & \textbf{99.72} & \textbf{93.80} & \textbf{98.80} & 84.35 & \textbf{94.04} & \textbf{86.97} & \textbf{96.38} \\
\bottomrule
\end{tabularx}
\label{tab: easycom_known} 
\end{table*}

\subsubsection{Implementations}
We implement our method in the PyTorch framework \cite{paszke2019pytorch}, and all experiments are run on an Amazon AWS server with eight A100 GPUs, each training session deployed on a single GPU. 
We re-implement the baseline methods SSiam \cite{sharma2019self}, CCL \cite{sharma2020clustering} and TSiam \cite{sharma2019self} in PyTorch for a fair comparison, based on the papers and the released code in Matlab. It takes around $7$ hours to run a full training session for our video-centralised transformer on a BBT video, while the total training time increases to approximately $23$ hours on an EasyCom-Clustering video. We evaluate all methods on each of the $6$ BBT videos (or episodes), and similarly on each of the $22$ EasyCom-Clustering videos (or sessions), while the overall performance is obtained by averaging the video-level performance on each of the two datasets, respectively.

\subsection{Results}
\subsubsection{Ablation studies on different components}
We first justify the design of our system framework through an ablation study on a single Easycom-Clustering video, namely \say{V04}, under the setting of known cluster numbers. When changing the design of one component, we freeze others for a fair comparison. A transformer 
trained with the contrastive learning illustrated in Fig. \ref{fig: idea_explain} (\textit{Middle}), i.e. CT TRSF., is used as the baseline model. 

We mainly examine four key components of our framework, 1). the introduction of video centres, 2). the temporal augmentation techniques $\op{\tau}$, 3). different methods of representing video centres, and 4). the learning rate $\eta$ for updating video centres.

The first component that we study is the introduction of video centres, mainly through the application of video-centralised loss in Eq. \ref{eq: VC loss}. 

The next ablation is placed on the temporal augmentation techniques, and we explore two different ways of implementing $\op{\tau}$, i.e. a). a uniform sampler that samples uniformly from a temporal interval, and b). a consecutive sampler as in Eq. \ref{eq:tau}. 

We also show the effectiveness of three different ways to represent video centres $\mathbf{c}_a$, which are: a). use the value of $\mathbf{z}_a^{\op{\tau}}$ from the last forward propagation as $\mathbf{c}_a$, b). the incomplete version of Algorithm \ref{alg:optimise} which skips step \ref{algo_step: re-compute-1}-\ref{algo_step: re-compute-2}, i.e. without re-computing video centres with Eq. \ref{eq: whole_track_center}, and c). the complete Algorithm \ref{alg:optimise}.

For the video centre's learning rate $\eta$, we explore two strategies, which are: a). the simple incremental learning rate as in \cite{yang2017towards}, and b). the learning rate that is proportional to the learning rate of $\boldsymbol{\theta}$, i.e. $\eta=p\xi$ where $p > 0$.

We consider the option \say{a} in each key component as the naive solution, and they are gradually replaced by other potential solutions to examine the introduced improvement. The ablation study results are shown in Table \ref{tab: ablation}. We can see that the initial performance of our video-centralised transformer (VC TRSF.) is not working well with all naive solutions embedded when compared with the baseline CT TRSF. However, the switching from the uniform sampler to the consecutive one as in Eq. \ref{eq:tau} has slightly enhanced the accuracy, while even higher improvement can be introduced via using the incomplete version of Algorithm \ref{alg:optimise} to represent and to optimise video centres. The complete Algorithm \ref{alg:optimise} can better cluster the videos, while keeping $\eta$ in proportional to $\xi$ can significantly 
improve the performance than a simple incremental $\eta$. 

\subsubsection{Ablation studies on Positional Embedding}
\label{sec:ablation_study_pe}
We also investigate whether applying the Positional Embedding (PE) strategy can further enhance the clustering performance of the proposed VC TRSF. 
Learnable PE is a prevalent component in video-based transformers like TimeSFormer \cite{bertasius2021space} or SVT \cite{ranasinghe2022self} to introduce awareness of temporal positions. 
However, video face clustering can be insensitive to temporal orderings, and we choose not to use PE in this work. 
To illustrate this decision, we compare the performance of our VC TRSF under three different options: 1). No PE, 2). the scaled PE \cite{li2019neural}, and 3). the interpolated PE \cite{bertasius2021space}, evaluated on the Easycom-Clustering video \say{V08} of known clusters.
We also integrate the performance of different baseline approaches on the same video. As demonstrated in Table \ref{tab: pe_ablation}, the performance of our VC TRSF. has surpassed that of baselines under all three options. 
Among them, applying Interpolated PE can gain better WCP than the rest two options. 
However, no PE can lead to the highest NMI at the cost of a slightly dropped WCP. 
Neither interpolated nor scaled PE can introduce a significant improvement compared to not using them, which verifies the insensitivity of video face clustering to temporal orderings. 
Following this result, we discard PE in this work towards a simplified training process and better computational efficiency. 

\subsubsection{Clustering Performance}
The performance of different methods on BBT and EasyCom-Clustering videos with known cluster numbers are shown in Table \ref{tab: bbt_known} and \ref{tab: easycom_known}, respectively. The simple video face clustering framework (Temp. AVG) \cite{Sharma2017ASimple} typically gives the lowest clustering performance, which is in line with the expectation since no learning process is involved. When compared with the Siamese-based-MLP methods (SSiam, CCL, TSiam), a transformer trained with the naive extension of contrastive loss (CT TRSF.) can already achieve competitive accuracy. With our video-centralised learning, the transformer (VC TRSF.) has outperformed all baselines on $5$ out of $6$ BBT videos and on $20$ out of $22$ EasyCom-Clustering videos in terms of WCP, which verifies the advantages of our video-centralised transformer-based representation.

\begin{table*}[t!]
\caption{Video clustering performance of different methods on BBT dataset with unknown cluster numbers. The true cluster number of each video is post-appended to the video name. \protect\say{\#C} refers to the predicted cluster number. 
}
\centering
\scriptsize
\begin{tabularx}{0.7\linewidth}{svivivivivivi}
\toprule
& \multicolumn{12}{c}{\textbf{BBT} (\%), Unknown Cluster Number} \\
\cmidrule(l{5pt}r{3pt}){2-13} 
\multirow{2}{*}{\textbf{Methods}} & \multicolumn{2}{c}{s01e01 (8)} & \multicolumn{2}{c}{s01e02 (6)} & \multicolumn{2}{c}{s01e03 (26)} & \multicolumn{2}{c}{s01e04 (28)} & \multicolumn{2}{c}{s01e05 (25)} & \multicolumn{2}{c}{s01e06 (37)} \\
 & \#C & NMI & \#C & NMI & \#C & NMI & \#C & NMI & \#C & NMI & \#C & NMI \\
 \cmidrule(l{16pt}r{16pt}){1-1} 
   \cmidrule(l{5pt}r{3pt}){2-3} 
  \cmidrule(l{5pt}r{3pt}){4-5}
  \cmidrule(l{5pt}r{3pt}){6-7}
  \cmidrule(l{5pt}r{3pt}){8-9}
  \cmidrule(l{5pt}r{3pt}){10-11}
  \cmidrule(l{5pt}r{3pt}){12-13} 
Temp. AVG \cite{Sharma2017ASimple} & 7 & 95.43 & 8 & 89.80 & 11 & 91.01 & 13 & 87.80 & 11 & \textbf{94.45} & 22 & 83.07 \\
SSiam \cite{sharma2019self} & \textbf{7} & 87.05 & 6 & 89.44 & 16 & 90.20 & 10 & 85.58 & 8 & 89.12 & 9 & 71.37 \\
CCL \cite{sharma2020clustering} & 6 & 96.45 & 5 & 98.55 & 13 & 93.54 & 12 & 89.60 & 7 & 89.15 & 12 & 79.53 \\
TSiam \cite{sharma2019self} & 6 & 95.59 & 5 & 98.55 & \textbf{18} & \textbf{95.24} & \textbf{14} & 90.83 & \textbf{14} & 93.66 & 12 & 82.62 \\
CT TRSF. & 5 & 94.80 & \textbf{6} & 97.32 & 11 & 90.81 & 11 & 88.84 & 9 & 89.25 & 12 & 79.20 \\
   \cmidrule(l{5pt}r{3pt}){2-3} 
  \cmidrule(l{5pt}r{3pt}){4-5}
  \cmidrule(l{5pt}r{3pt}){6-7}
  \cmidrule(l{5pt}r{3pt}){8-9}
  \cmidrule(l{5pt}r{3pt}){10-11}
  \cmidrule(l{5pt}r{3pt}){12-13} 
VC TRSF. (Ours) & 6 & \textbf{97.06} & 5 & \textbf{98.81} & 12 & 94.11 & 11 & \textbf{91.03} & 11 & 92.50 & \textbf{29} & \textbf{84.17} \\
\bottomrule
\end{tabularx}
\label{tab: bbt_unknown} 
\end{table*}
\begin{table*}[ht]
\caption{Video clustering performance of different methods on EasyCom-Clustering dataset with unknown cluster numbers. The true cluster number of each video is post-appended to the video name. \protect\say{\#C} refers to the predicted cluster number. } 
\centering
\scriptsize
\begin{tabularx}{0.99\linewidth}{svivivivivivivivivivivi}
\toprule
& \multicolumn{22}{c}{\textbf{EasyCom-Clustering} (\%), Unknown Cluster Number} \\
\cmidrule(l{5pt}){2-23} 
\multirow{2}{*}{\textbf{Methods}} & \multicolumn{2}{c}{V01 (4)} & \multicolumn{2}{c}{V02 (3)} & \multicolumn{2}{c}{V03 (3)} & \multicolumn{2}{c}{V04 (3)} & \multicolumn{2}{c}{V05 (4)} & \multicolumn{2}{c}{V06 (3)} & \multicolumn{2}{c}{V07 (3)} & \multicolumn{2}{c}{V08 (4)} & \multicolumn{2}{c}{V09 (3)} & \multicolumn{2}{c}{V10 (3)} & \multicolumn{2}{c}{V11 (3)} \\
 & \#C & NMI & \#C & NMI & \#C & NMI & \#C & NMI & \#C & NMI & \#C & NMI & \#C & NMI & \#C & NMI & \#C & NMI & \#C & NMI & \#C & NMI \\
 \cmidrule(l{9pt}r{9pt}){1-1} 
  \cmidrule(l{6pt}r{0pt}){2-3}  \cmidrule(l{6pt}r{0pt}){4-5}  \cmidrule(l{6pt}r{0pt}){6-7}
  \cmidrule(l{6pt}r{0pt}){8-9}
   \cmidrule(l{6pt}r{0pt}){10-11}
    \cmidrule(l{6pt}r{0pt}){12-13}
    \cmidrule(l{6pt}r{0pt}){14-15}
    \cmidrule(l{6pt}r{0pt}){16-17}
    \cmidrule(l{6pt}r{0pt}){18-19}
    \cmidrule(l{6pt}r{0pt}){20-21}
    \cmidrule(l{6pt}r{0pt}){22-23}
Temp. AVG \cite{Sharma2017ASimple} & 2 & 61.36 & 2 & 19.46 & 2 & 55.09 & 2 & 69.27 & 3 & 51.58 & 3 & 37.56 & 3 & 81.04 & \textbf{3} & 59.44 & 2 & 84.08 & 2 & 18.82 & 1 & 0.00 \\
SSiam \cite{sharma2019self} & 5 & 92.15 & 4 & 76.76 & 4 & 85.80 & 5 & 82.86 & 5 & 82.71 & 4 & 77.71 & 4 & 90.05 & 6 & \textbf{84.54} & 4 & 89.89 & 5 & 81.56 & 4 & 78.57 \\
CCL \cite{sharma2020clustering} & 7 & 90.05 & 9 & 71.92 & 7 & 86.21 & 4 & 83.16 & 10 & 84.37 & 7 & 74.60 & 5 & 88.21 & 9 & 84.53 & 5 & 91.92 & 6 & 78.33 & 7 & 78.73 \\
TSiam \cite{sharma2019self} & 7 & \textbf{92.31} & 7 & 80.12 & 7 & 85.09 & 7 & 82.16 & 8 & \textbf{86.34} & 7 & \textbf{84.04} & 7 & 88.17 & 7 & 83.86 & 5 & 86.06 & 6 & 83.11 & 6 & 81.09 \\
CT TRSF. & \textbf{4} & 90.16 & 3 & 74.41 & 3 & 94.40 & 3 & 85.2 & 5 & 82.37 & 2 & 33.74 & 3 & 91.54 & 6 & 79.45 & 3 & 85.99 & 4 & 70.40 & 3 & 80.37 \\
   \cmidrule(l{6pt}r{0pt}){2-3}  \cmidrule(l{6pt}r{0pt}){4-5}  \cmidrule(l{6pt}r{0pt}){6-7}
  \cmidrule(l{6pt}r{0pt}){8-9}
   \cmidrule(l{6pt}r{0pt}){10-11}
    \cmidrule(l{6pt}r{0pt}){12-13}
    \cmidrule(l{6pt}r{0pt}){14-15}
    \cmidrule(l{6pt}r{0pt}){16-17}
    \cmidrule(l{6pt}r{0pt}){18-19}
    \cmidrule(l{6pt}r{0pt}){20-21}
    \cmidrule(l{6pt}r{0pt}){22-23}
VC TRSF. (Ours) & 3 & 90.23 & \textbf{3} & \textbf{82.25} & \textbf{3} & \textbf{94.86} & \textbf{3} & \textbf{90.69} & \textbf{3} & 81.82 & \textbf{3} & 79.81 & \textbf{3} & \textbf{93.29} & 6 & 76.73 & \textbf{3} & \textbf{98.17} & \textbf{3} & \textbf{89.47} & \textbf{3} & \textbf{87.12} \\
 \midrule \midrule 
& \multicolumn{2}{c}{V12 (3)} & \multicolumn{2}{c}{V13 (3)} & \multicolumn{2}{c}{V14 (5)} & \multicolumn{2}{c}{V15 (3)} & \multicolumn{2}{c}{V16 (3)} & \multicolumn{2}{c}{V17 (3)} & \multicolumn{2}{c}{V18 (3)} & \multicolumn{2}{c}{V19 (3)} & \multicolumn{2}{c}{V20 (3)} & \multicolumn{2}{c}{V21 (5)} & \multicolumn{2}{c}{V22 (4)} \\
 & \#C & NMI & \#C & NMI & \#C & NMI & \#C & NMI & \#C & NMI & \#C & NMI & \#C & NMI & \#C & NMI & \#C & NMI & \#C & NMI & \#C & NMI \\
   \cmidrule(l{6pt}r{0pt}){2-3}  \cmidrule(l{6pt}r{0pt}){4-5}  \cmidrule(l{6pt}r{0pt}){6-7}
  \cmidrule(l{6pt}r{0pt}){8-9}
   \cmidrule(l{6pt}r{0pt}){10-11}
    \cmidrule(l{6pt}r{0pt}){12-13}
    \cmidrule(l{6pt}r{0pt}){14-15}
    \cmidrule(l{6pt}r{0pt}){16-17}
    \cmidrule(l{6pt}r{0pt}){18-19}
    \cmidrule(l{6pt}r{0pt}){20-21}
    \cmidrule(l{6pt}r{0pt}){22-23}
Temp. AVG \cite{Sharma2017ASimple} & 2 & 48.56 & 3 & 39.30 & 2 & 52.23 & 1 & 0.00 & 2 & 46.91 & 2 & 49.97 & 2 & 46.79 & 2 & 65.42 & 3 & 77.97 & 2 & 34.74 & 3 & 59.21 \\
SSiam \cite{sharma2019self} & 4 & 82.46 & 6 & 86.50 & \textbf{7} & 83.37 & 3 & 79.52 & 3 & 92.41 & 3 & 62.06 & 5 & 90.18 & 3 & 94.06 & 4 & 88.08 & 8 & 81.82 & \textbf{5} & 82.60 \\
CCL \cite{sharma2020clustering} & 6 & 86.59 & 9 & 86.59 & 11 & 88.12 & 5 & 83.43 & 6 & 89.31 & 5 & 85.38 & 7 & 84.09 & 5 & 88.61 & 6 & 89.92 & 9 & 84.14 & 11 & 83.15 \\
TSiam \cite{sharma2019self} & 5 & 82.01 & 10 & 77.91 & 11 & \textbf{91.48} & 7 & \textbf{84.81} & 5 & 94.69 & 3 & \textbf{91.00} & 6 & 84.57 & 4 & 92.26 & 6 & 85.38 & 7 & \textbf{84.20} & 12 & \textbf{84.82} \\
CT TRSF. & 3 & 81.08 & 6 & 68.00 & 8 & 81.03 & \textbf{3} & 83.99 & 3 & 83.60 & \textbf{3} & 73.13 & 4 & 81.11 & 3 & 93.41 & \textbf{3} & \textbf{90.97} & \textbf{5} & 80.42 & 7 & 70.37 \\
   \cmidrule(l{6pt}r{0pt}){2-3}  \cmidrule(l{6pt}r{0pt}){4-5}  \cmidrule(l{6pt}r{0pt}){6-7}
  \cmidrule(l{6pt}r{0pt}){8-9}
   \cmidrule(l{6pt}r{0pt}){10-11}
    \cmidrule(l{6pt}r{0pt}){12-13}
    \cmidrule(l{6pt}r{0pt}){14-15}
    \cmidrule(l{6pt}r{0pt}){16-17}
    \cmidrule(l{6pt}r{0pt}){18-19}
    \cmidrule(l{6pt}r{0pt}){20-21}
    \cmidrule(l{6pt}r{0pt}){22-23}
VC TRSF. (Ours) & \textbf{3} & \textbf{90.89} & \textbf{3} & \textbf{92.99} & 8 & 84.50 & 2 & 72.89 & \textbf{3} & \textbf{97.22} & 4 & 86.23 & \textbf{3} & \textbf{93.15} & \textbf{3} & \textbf{98.41} & 4 & 89.05 & 7 & 79.57 & 7 & 74.27 \\
\bottomrule
\end{tabularx}
\label{tab: easycom_unknown} 
\end{table*}

The evaluation results on videos with unknown cluster number are demonstrated in Table \ref{tab: bbt_unknown} (BBT) and \ref{tab: easycom_unknown} (EasyCom-Clustering), respectively. Different from the evaluation with known cluster numbers, we focus on the accuracy of the predicted cluster number and the achieved NMI. As can be seen, although TSiam is serving a competitive baseline, the best NMIs are still obtained by our VC TRSF. on $4$ out of all $6$ BBT videos, and on $12$ out of all $22$ EasyCom-Clustering videos. 

\begin{table*}[t!]
\caption{The overall performance of different methods on BBT and EasyCom-Clustering with cluster number known/unknown. \protect\say{\#C DIF} refers to the sum of the absolute differences between the predicted and true cluster number of all videos of the dataset, which is the smaller the better. }
\centering
\scriptsize
\begin{tabularx}{0.8\linewidth}{kyyyyyyyy}
\toprule
 & \multicolumn{4}{c}{\textbf{BBT}} & \multicolumn{4}{c}{\textbf{EasyCom-Clustering}} \\
 \cmidrule(l{10pt}r{10pt}){2-5}  \cmidrule(l{10pt}r{10pt}){6-9} 
\multicolumn{1}{c}{\multirow{2}{*}{\textbf{Methods}}} & \multicolumn{2}{c}{Known} & \multicolumn{2}{c}{Unknown} & \multicolumn{2}{c}{Known} & \multicolumn{2}{c}{Unknown} \\
& NMI (\%) & WCP (\%) & \#C DIF & NMI (\%) & NMI (\%) & WCP (\%) & \#C DIF & NMI (\%) \\
 \cmidrule(l{12pt}r{12pt}){1-1} 
 \cmidrule(l{8pt}r{8pt}){2-3} 
  \cmidrule(l{10pt}r{8pt}){4-5}
   \cmidrule(l{8pt}r{8pt}){6-7}
    \cmidrule(l{10pt}r{8pt}){8-9} 
Temp. AVG \cite{Sharma2017ASimple} & 88.57 & 95.84 & 62 & 90.26 & 65.05 & 78.44 & 25 & 48.13 \\
SSiam \cite{sharma2019self} & 92.34 & 97.01 & 74 & 85.46 & 83.42 & 93.07 & 27 & 83.89 \\
CCL \cite{sharma2020clustering} & 93.98 & 97.11 & 75 & 91.14 & 85.84 & 94.62 & 82 & 84.61 \\
TSiam \cite{sharma2019self} & 94.11 & 97.04 & 61 & 92.75 & 87.10 & 95.84 & 76 & 85.70 \\
CT TRSF. & 91.86 & 96.77 & 76 & 90.04 & 84.48 & 94.49 & 15 & 79.78 \\
 \cmidrule(l{8pt}r{8pt}){2-3} 
  \cmidrule(l{10pt}r{8pt}){4-5}
   \cmidrule(l{8pt}r{8pt}){6-7}
    \cmidrule(l{10pt}r{8pt}){8-9} 
VC TRSF. (Ours) & \textbf{94.20} & \textbf{97.50} & \textbf{56} & \textbf{92.95} & \textbf{91.01} & \textbf{97.45} & \textbf{15} & \textbf{87.44}  \\
\bottomrule
\end{tabularx}
\label{tab: overall_res} 
\end{table*}

The overall performance on two datasets under known/unknown clustering settings is exhibited in Table \ref{tab: overall_res}. The overall accuracy achieved by Temp. AVG \cite{Sharma2017ASimple} is lower than other learning-based models. 
Among all Siamese-MLP-based methods, TSiam is the most competitive one that surpasses the performance of others on both datasets and two clustering scenarios. CT TRSF., on the other hand, has shown comparable clustering results, which is already impressive considering that no previous works have ever examined the potential of a self-supervised transformer to obtain video-level representation for clustering. 
Our VC TRSF., however, have exceeded the performance of all other approaches on those two datasets.
Such superior performance verifies our method's generality to various ConvNet backbones, since the facial embeddings of two datasets are extracted with two face models of significantly different architectures, pre-trained datasets, and learning losses, respectively.  
It can also be observed that the improvement achieved on the EasyCom-Clustering dataset is generally more significant than on BBT videos. This is not surprising, considering that each BBT video only consists of a small amount of data (around $650$ tracks with $38$ thousand facial images). For now, most transformers \cite{yuan2021tokens,touvron2021training,touvron2021going} are trained on large-scale datasets, such as ImageNet \cite{deng2009imagenet} with more than 1.3 million training images, to achieve performance comparable to state-of-the-art. 
Our results on BBT videos demonstrate that the transformer can also be used to perform video understanding with small-scale data in a self-supervised fashion. 
The more significant improvement achieved on EasyCom-Clustering videos (around $4\,270$ tracks with $73$ thousand facial images per video) illustrates that transformers can still benefit from larger-scale data to fulfil the potential better. 
Whatever the situation, the proposed video-centralised learning plays a critical role in guiding the transformer towards a more discriminate and more centralised video-level representation. 

\begin{figure*}[ht!]
\centering
\includegraphics[width=0.99\linewidth]{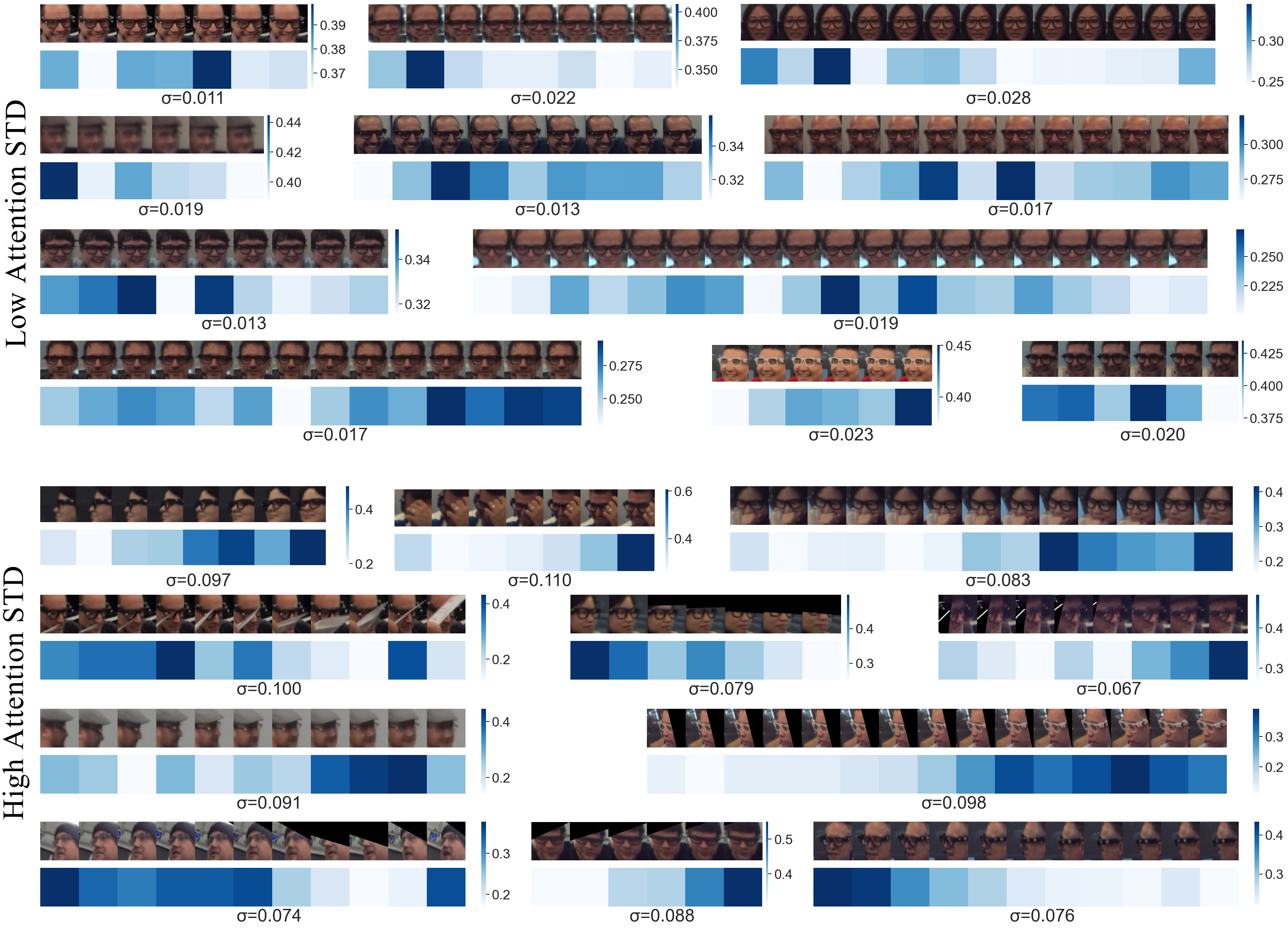} 
\caption{The visualisation of temporal attention score predicted by our video-centralised transformer on EasyCom-Clustering face tracks, which are organised by the standard deviation (STD) $\sigma$ of their attention scores.
A higher $\sigma$ intuitively exhibits that the transformer attends each frame more distinguishably, and a lower $\sigma$ suggest fewer needs to treat each time step differently.
The deeper the heatmap, the more attention is paid to that frame.  
(Best seen in colour)
}
\label{fig: temporal_attention}
\end{figure*}

\subsubsection{Advantages of transformers}
We also investigate why the video-level representation obtained by our VC TRSF. can outperform others through an intuitive visualisation. 
In Fig. \ref{fig: temporal_attention},
our VC TRSF.'s attention score \cite{vig2019multiscale} on the temporal dimension of several EasyCom-Clustering face tracks are portrayed.
The heatmap under each face track stands for the L2-Normalised attention scores,
and the deeper it is, the more attentive is in that position. 
We also compute the standard deviation (STD) $\sigma$ of attention scores for each track, organising them into two categories of low and high $\sigma$ values.
Intuitively, a lower $\sigma$ suggests that the transformer tends to weigh each frame more homogenously, and a higher $\sigma$ exhibits that the model attends each frame more distinguishably.
As can be discovered from Fig. \ref{fig: temporal_attention}, face tracks with lower $\sigma$ values (e.g., $\sigma<0.025$) demonstrate generally fewer changes across the temporal dimensional, and such trends are successfully captured by the transformer assigning temporal attention scores of fewer variances. 
As for those tracks with higher attention STDs (e.g., $\sigma>0.065$), we can observe the presence of potential visual distractors in most cases. 
Those distractors, e.g. face with a certain amount of occlusions, profile faces that are more difficult to be recognised, images taken under extremely dark illuminations, faces that are partially outside of the frame (those with partially invisible contents), etc., are mostly identified with lower attentive scores on those frames. Those images with better visual qualities are generally granted a higher level of attention to better contribute to generating video-level representations. 
This kind of temporal awareness is a unique advantage of the transformer model and cannot be achieved in previous works such as TSiam \cite{sharma2019self} or CCL \cite{sharma2020clustering}, and so on. 

\section{Conclusion}
In this paper, we demonstrate a self-supervised transformer along with the video-centralised learning to address the video face clustering problem. The transformer is employed to obtain the video-level representation for each face track, which is a valuable perspective untouched by previous works. To improve the self-supervised learning for videos, we propose the video-centralised learning to train the transformer, motivated by the idea that each face track should be assigned with a distinct video centre in the latent space. 
We also release a new dataset, dubbed EasyCom-Clustering, which is the first large-scale video face clustering dataset on egocentric videos.
We evaluate the performance of the proposed video-centralised transformer with several state-of-the-art methods, and the superior performance achieved on two datasets has verified its effectiveness. The experimental results also reveal that the transformer can be applied to a self-supervised video understanding task with small-scale data and can achieve state-of-the-art performance. The proposed video-centralised learning can be inspiring to future research on self-supervised learning and transformer in video-based vision problems.
\ifCLASSOPTIONcaptionsoff
  \newpage
\fi

\section*{Acknowledgements}
The work of Mingzhi Dong was supported by China Postdoctoral Science Foundation under Grant No. 2022M720767.

\bibliographystyle{IEEEtran}
\bibliography{ref.bib}
\begin{IEEEbiography}[{\includegraphics[width=1in,height=1.25in,clip,keepaspectratio]{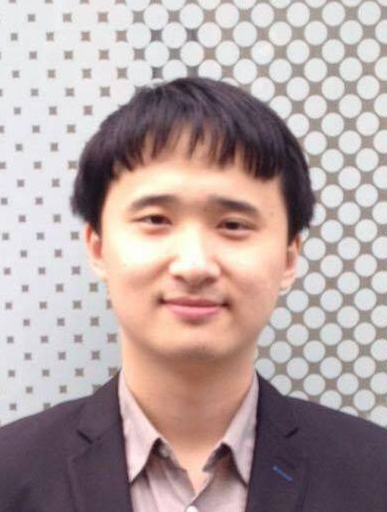}}]{Yujiang Wang}
received his PhD degree from Imperial College London on February 2021, and he is currently working as a Postdoctoral Researcher at University of Oxford. He obtained a BSc degree in Architecture from Tsinghua University in 2010, and he was awarded two MSc with Distinction from University College London and Imperial College London, respectively. His research interest centres around video face clustering, lip-reading, smart wearable devices, healthcare AI, etc. 
\end{IEEEbiography}
 
\begin{IEEEbiography} 
[{\includegraphics[width=1in,height=1.25in,clip,keepaspectratio]{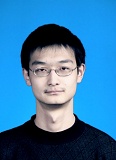}}]
{Mingzhi Dong} 
received B.Eng. degree in 2010 and M.Eng. degree in 2013, both from Beijing University of Posts and Telecommunications. He received Ph. D degree from University College London in 2019. Currently, he is a postdoctoral research fellow at Fudan University. His research interests include metric learning, reinforcement learning, etc.
\end{IEEEbiography}
 
\begin{IEEEbiography}
 [{\includegraphics[width=1in,height=1.25in,clip,keepaspectratio]{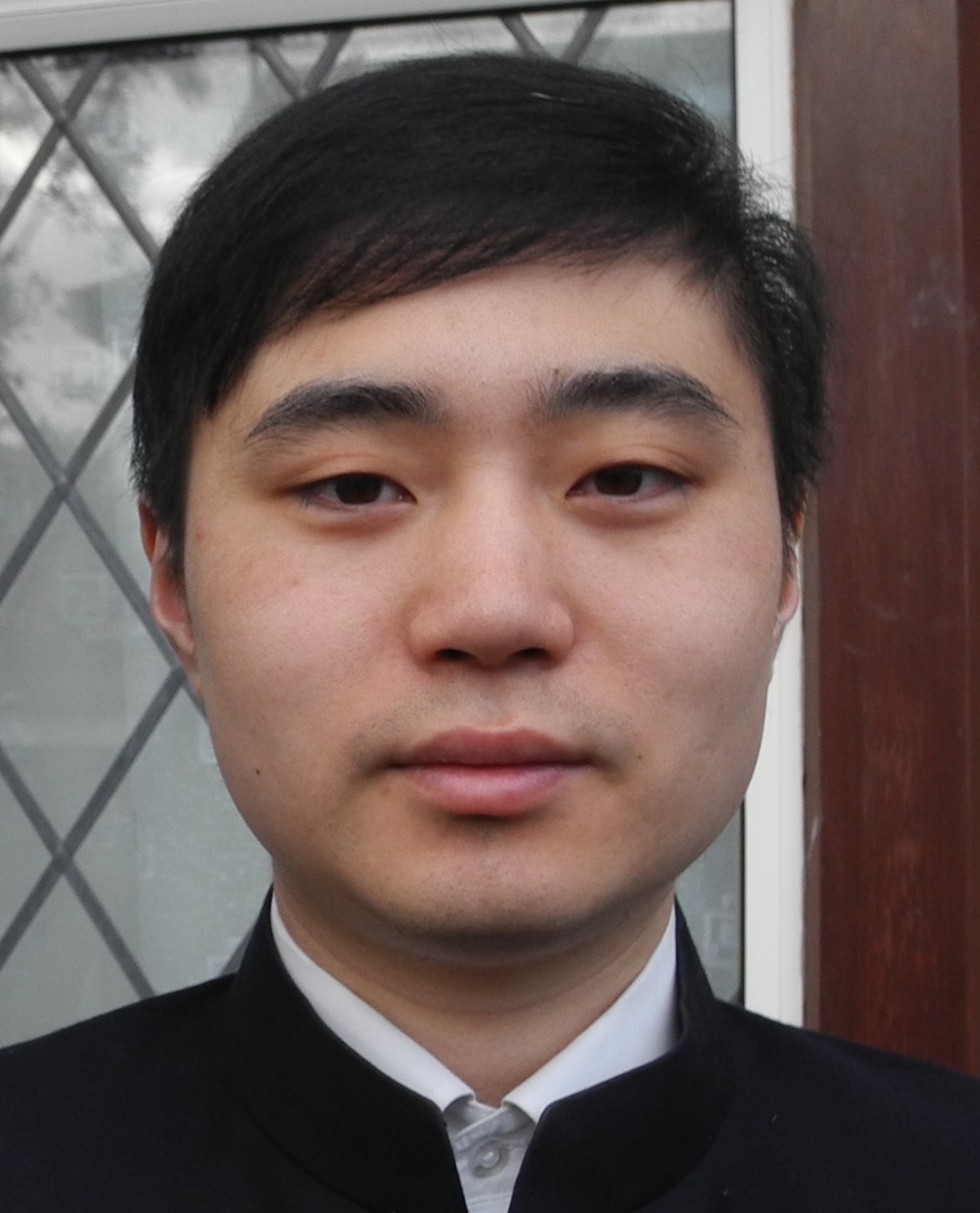}}]
 {Jie Shen} is a scientific research manager at Meta AI and an honorary research fellow at the Department of Computing at Imperial College London. He received his B.Eng. in electronic engineering from
Zhejiang University in 2005, and his MSc in advanced computing and Ph.D. from Imperial College London in 2008 and 2014. His research interests include facial analysis, computer vision, affective computing, and social robots. He is a member of the IEEE.
\end{IEEEbiography}
 
\begin{IEEEbiography} 
[{\includegraphics[width=1in,height=1.25in,clip,keepaspectratio]{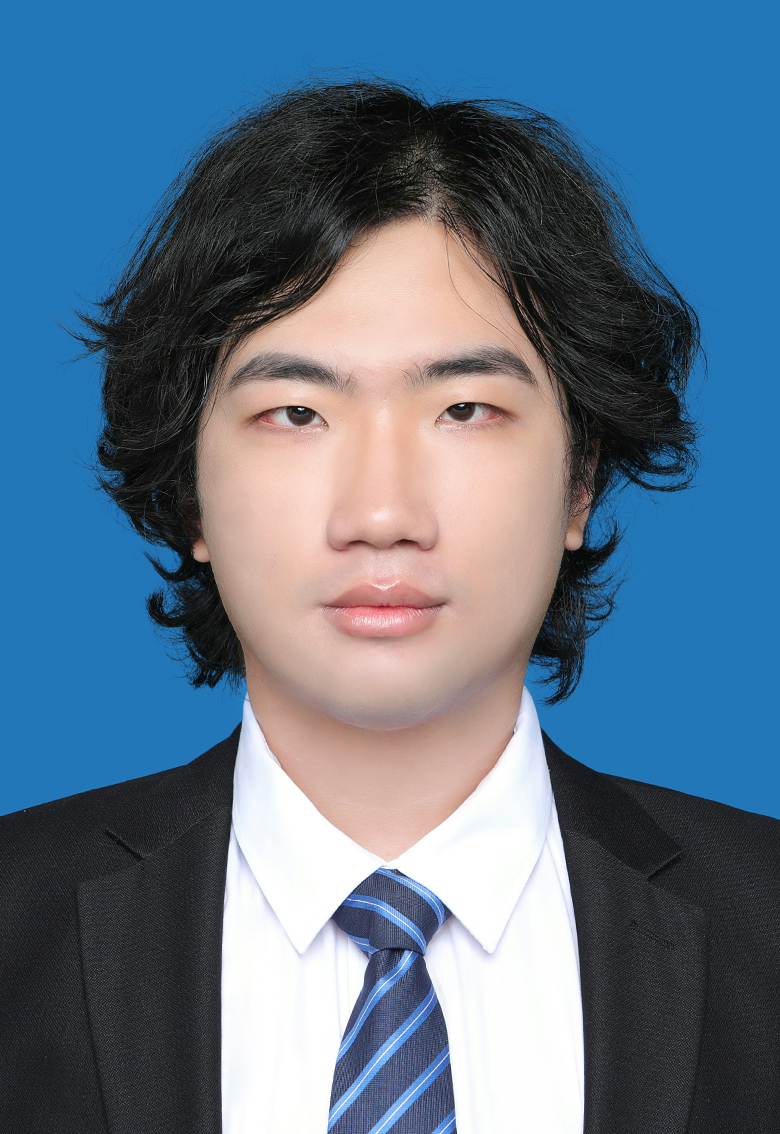}}]
{Yiming Luo} is currently a postgraduate student at Imperial College London. He received an B.Eng. Degree in 2022 at Tsinghua University, China. His research interests include affective computing, gesture generating and human-computer interaction, etc.
\end{IEEEbiography}
 
\begin{IEEEbiography} 
 [{\includegraphics[width=1in,height=1.25in,clip,keepaspectratio]{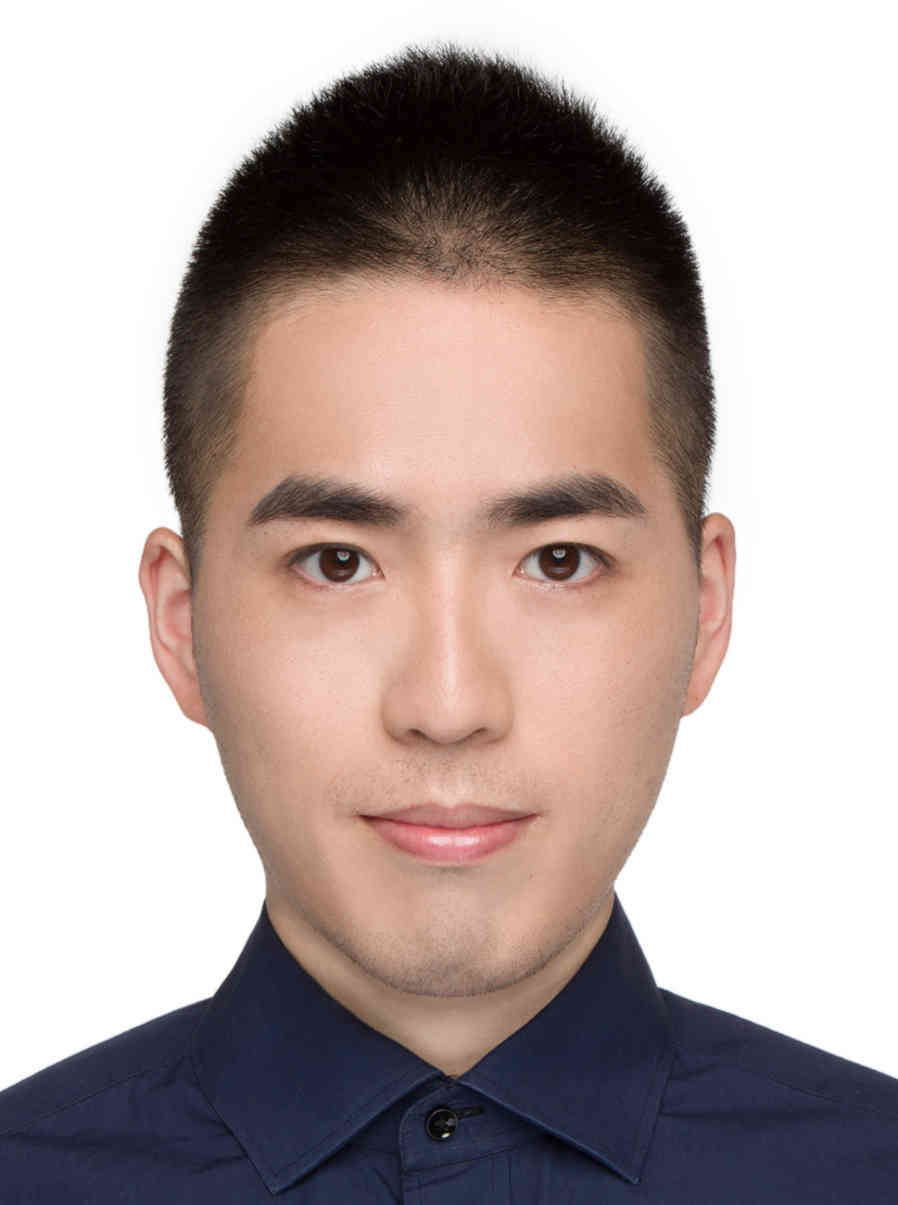}}]
{Yiming Lin} is a research scientist at Meta AI. He received his PhD degree in 2021 and MSc degree in 2016 from Imperial College London. His research interests include generative models, machine learning robustness and fairness, face parsing, face recognition and facial attribute analysis. He is a member of IEEE.
\end{IEEEbiography}
 
\begin{IEEEbiography} 
 [{\includegraphics[width=1in,height=1.25in,clip,keepaspectratio]{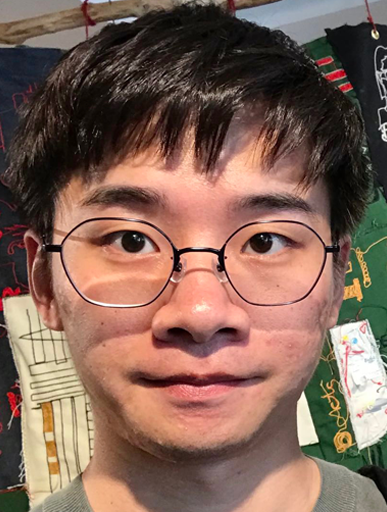}}]
{Pingchuan Ma}
received his BSc degree from Beihang University in 2015 and his MSc degree and PhD degree from Imperial College London in 2017 and 2022, respectively. Currently, he is a postdoctoral researcher at Meta AI. His research interest centres around audiovisual speech recognition and self-supervised learning. He is a member of the IEEE.
\end{IEEEbiography}

\begin{IEEEbiography} 
 [{\includegraphics[width=1in,height=1.25in,clip,keepaspectratio]{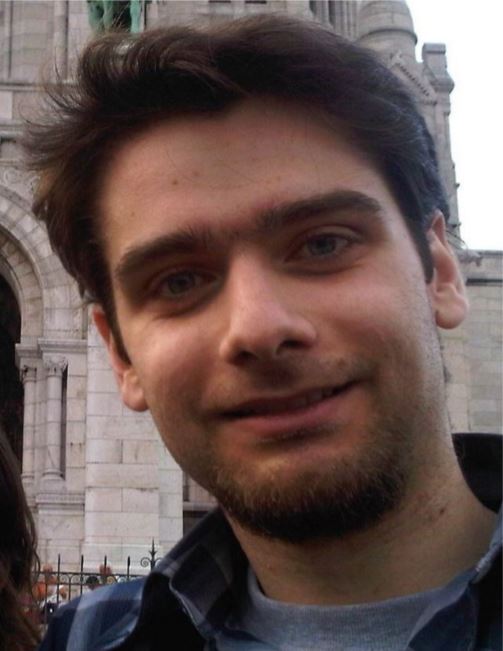}}]
{Stavros Petridis} is a scientific research manager at Meta AI and an honarary research fellow at the Department of Computing at Imperial College London. He received his BSc degree in electrical and computer engineering from the Aristotle University of Thessaloniki, Greece in 2004 and his MSc degree in Advanced Computing and Ph.D. from Imperial College London in 2005 and 2012, respectively. He has been a research intern in the Image Processing Group at University College London and the Field Robotics Centre, Robotics Institute, Carnegie Mellon University and a visiting researcher at the affect analysis group at University of Pittsburgh. His research interests lie in the areas of pattern recognition and machine learning and their application to multimodal recognition of human behaviour. He is a member of the IEEE.
\end{IEEEbiography}

\begin{IEEEbiography} 
 [{\includegraphics[width=1in,height=1.25in,clip,keepaspectratio]{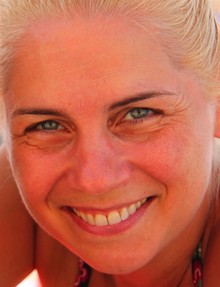}}]
 {Maja Pantic} is a professor in affective and behavioural computing in the Department of Computing at Imperial College London, UK. She was
the Research Director of Samsung AI Centre, Cambridge, UK from 2018 to 2020 and is currently an AI Scientific Research Lead at Meta London. She currently serves as an associate editor for both the IEEE Transactions on Pattern Analysis and Machine Intelligence and the IEEE Transactions on Affective Computing. She has received various awards for her work on automatic analysis of human behaviour,including the Roger Needham Award 2011. She is a fellow of the UK’s Royal Academy of Engineering, the IEEE, and the IAPR.
\end{IEEEbiography}
\vfill

\end{document}